\documentclass[10pt,journal]{IEEEtran}
\usepackage{amsmath,amsfonts}
 \usepackage[ruled,vlined]{algorithm2e}
\usepackage{array}
\usepackage{url}
\usepackage{graphicx}
\usepackage{multirow}
\usepackage{amssymb}
\usepackage[colorlinks=true]{hyperref}
\hypersetup{colorlinks, breaklinks, citecolor=[rgb]{0,0,1},}
\usepackage{soul}
\usepackage[table,xcdraw]{xcolor}
\definecolor{}{HTML}{2F54C8} 
\usepackage[normalem]{ulem}

\begin{document}

\title{Advancing Weakly-Supervised Change Detection in Satellite Images via Adversarial Class Prompting}

\author{Zhenghui Zhao,
Chen Wu,~\IEEEmembership{Member,~IEEE},
Di Wang,
Hongruixuan Chen,
Cuiqun Chen,\\
Zhuo Zheng, 
Bo Du,~\IEEEmembership{Senior Member,~IEEE},
Liangpei Zhang,~\IEEEmembership{Fellow,~IEEE}

 \thanks{This work is partly supported by the National Natural Science Foundation of China under Grant T2122014, and partly by the National Key Research and Development Program of China under Grant 2022YFB3903300. The numerical calculations in this paper have been done on the supercomputing system in the Supercomputing Center of Wuhan University.
 }
\thanks{Zhenghui Zhao, Chen Wu, and Liangpei Zhang are with the State Key Laboratory of Information Engineering in Surveying, Mapping and Remote Sensing, Wuhan University, China.  (\textit{Corresponding author: Chen Wu}.)

Di Wang and Bo Du are with the School of Computer Science, the National Engineering Research Center for Multimedia Software, the Institute of Artificial Intelligence, and the Hubei Key Laboratory of Multimedia and Network Communication Engineering, Wuhan University, China.

Cuiqun Chen is with the School of
Computer Science and Technology, Anhui University, China.

Hongruixuan Chen is with the Graduate School of Frontier Sciences, University of Tokyo, Japan, and the Institute of Geodesy and Photogrammetry, ETH Zürich, Switzerland.

Zhuo Zheng is with the Department of Computer Science, Stanford University, USA.



}}

\markboth{Journal of \LaTeX\ Class Files,~Vol.~14, No.~8, August~2021}%
{Shell \MakeLowercase{\textit{et al.}}: A Sample Article Using IEEEtran.cls for IEEE Journals}


\maketitle
\begin{abstract}
Weakly-Supervised Change Detection (WSCD) aims to distinguish specific object changes (e.g., objects appearing or disappearing) from background variations (e.g., environmental changes due to light, weather, or seasonal shifts) in paired satellite images, relying only on paired image (i.e., image-level) classification labels. This technique significantly reduces the need for dense annotations required in fully-supervised change detection. However, as image-level supervision only indicates whether objects have changed in a scene, WSCD methods often misclassify background variations as object changes, especially in complex remote-sensing scenarios. In this work, we propose an \underline{Adv}ersarial \underline{C}lass \underline{P}rompting (AdvCP) method to address this co-occurring noise problem, including two phases:
a) Adversarial Prompt Mining:  After each training iteration, we introduce adversarial prompting perturbations, using incorrect one-hot image-level labels to activate erroneous feature mappings. This process reveals co-occurring adversarial samples under weak supervision, namely background variation features that are likely to be misclassified as object changes.
b) Adversarial Sample Rectification: We integrate these adversarially prompt-activated pixel samples into training by constructing an online global prototype. This prototype is built from an exponentially weighted moving average of the current batch and all historical training data. Serving as an unbiased anchor, the global prototype guides the rectification of adversarial pixel samples.
Our AdvCP can be seamlessly integrated into current WSCD methods without adding additional inference cost. Experiments on ConvNet, Transformer, and Segment Anything Model (SAM)-based baselines demonstrate significant performance enhancements, achieving up to 7.37\%, 7.46\%, and 6.56\% IoU improvements on the WHU-CD, LEVIR-CD, and DSIFN-CD datasets. Furthermore, we demonstrate the generalizability of AdvCP to other multi-class weakly-supervised dense prediction scenarios. Code is available at \url{https://github.com/zhenghuizhao/AdvCP}.
\end{abstract}
 

\section{Introduction}\label{sec:intro}
\IEEEPARstart{C}{hange} Detection (CD) identifies specific object changes at the pixel level in multi-temporal remote sensing images, captured at different times but in the same scene. CD has applications in a wide range of domains, including environmental monitoring \cite{9wu,song2014remote}, urban planning \cite{Singh2020,lee2021local}, agriculture \cite{Jiang2022,Bai2023}, disaster assessment \cite{Tsvetkov2023,zheng2021building}, land use statistics \cite{Fatima2021}, and wildlife conservation \cite{Nawab2022,Shafique2022}.

\par With the increasing availability of high-resolution satellite images, the complexity of labeling has grown, making the cost of detailed pixel-level annotations a significant concern for their effective utilization. To reduce the expense of detailed pixel-level annotations, recent research has focused on Weakly-Supervised Change Detection (WSCD), shifting away from fully-supervised approaches \cite{wu2023,Huang2023}. Among various weakly-supervised paradigms, paired-image (\textit{i.e.}, image-level) classification labels have emerged as a cost-effective alternative, requiring the least amount of annotation effort \cite{zheng2021change, li2024ms, cheng2023out}.  In this work, we focus specifically on image-level WSCD, which we refer to simply as WSCD.

\par WSCD typically aims to obtain predictions of changes at the pixel level from a paired-image classification model. The prevailing approach to WSCD follows a three-step pipeline: 1) Paired remote-sensing images and their classification labels, indicating whether objects have `changed' or `unchanged', are fed into classification models. 2) Change localization maps (\textit{i.e.}, attention heatmaps) are extracted from these classification models using techniques such as Class Activation Maps (CAMs) \cite{zhou2016learning}. 3) Finally, the most salient pixels in the change localization maps are retained as pixel-level change predictions.

\begin{figure}[t]
	\centering
	\includegraphics[width=0.9999\linewidth]{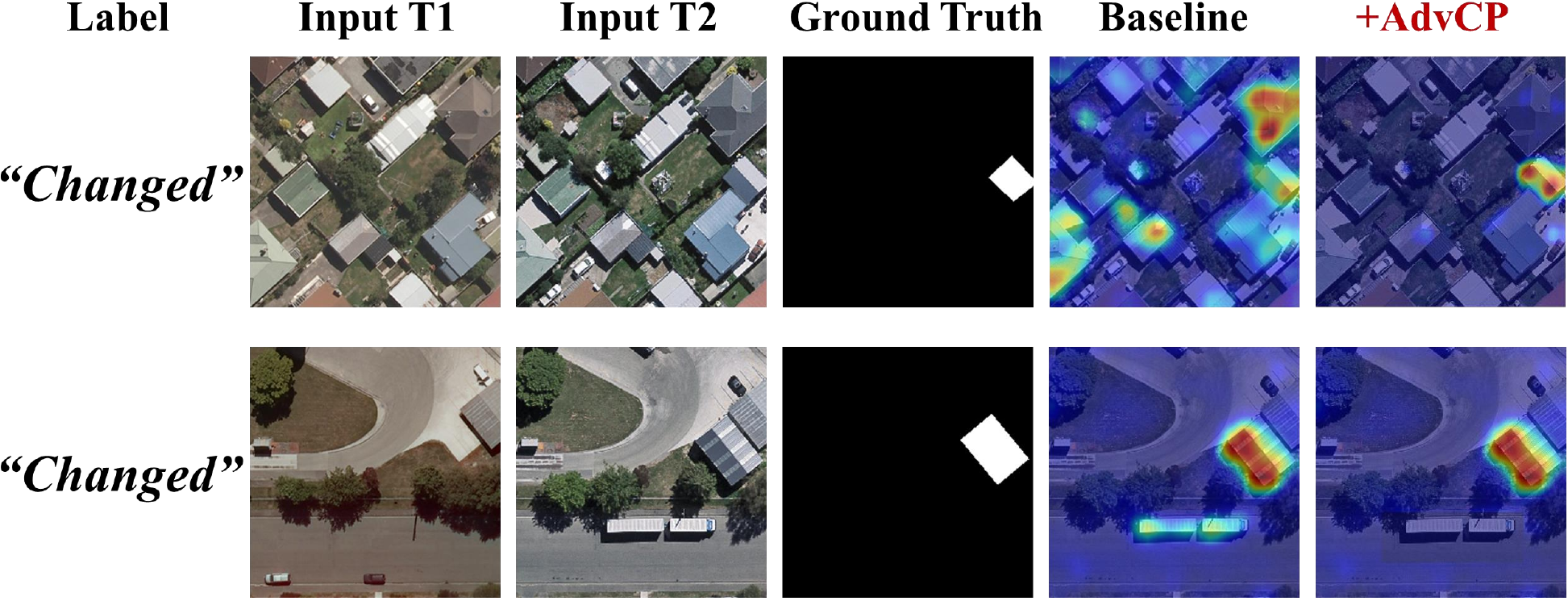}
	\caption{Problem of co-occurring noise from background variations under weak supervision. For example, in the second row of a SAM-based baseline, large trucks are mistakenly identified as changes in building change detection. After integrating our AdvCP method, the accuracy of the WSCD baseline is significantly improved. }
	\label{introduction1}
\end{figure}

Despite the structured pipeline, WSCD approaches often struggle to distinguish object changes from co-occurring background variations. The challenge arises because scenes labeled as `changed' typically contain both target object changes and unchanged background variations. These background variations frequently co-occur with object changes, making them difficult to isolate under image-level weak supervision. This co-occurring issue is particularly evident in remote sensing scenarios, with significant background noise caused by variations in lighting, weather, seasons, and human activities \cite{wu2024unet}. 

As observed in Fig. \ref{introduction1}, existing WSCD methods, even those based on powerful foundation models like the Segment Anything Model (SAM) \cite{kirillov2023segment}, fail to effectively address co-occurring noise due to the inherent limitations of image-level information. The co-occurring problem has not been explored in WSCD, while other weakly-supervised dense prediction methods often adopt complex approaches to tackle similar challenges \cite{lee2022weakly}.

\begin{figure}
	\centering
	\includegraphics[width=0.99\linewidth]{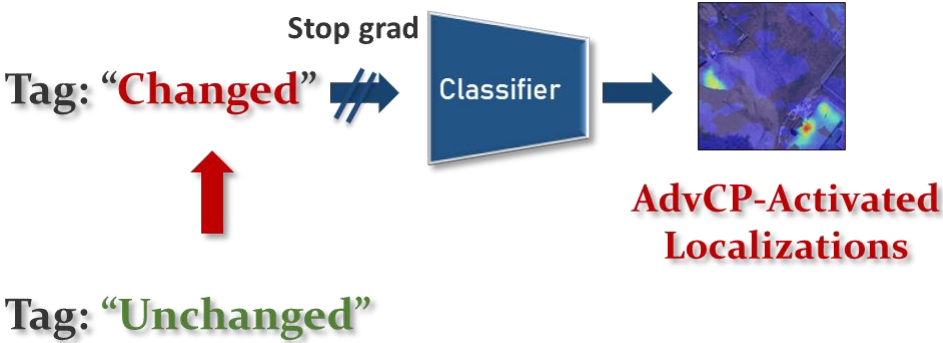}
	\caption{Conceptual pipeline of our AdvCP. Our AdvCP inverts image-level classification labels from `unchanged' to `changed' into the classifier, introducing adversarial prompt perturbations in unchanged scene images. This process acquires the corresponding AdvCP-activated localizations, \textit{i.e.}, background noise samples.}
	\label{introduction2}
\end{figure}

In this work, we propose Adversarial Class Prompting (AdvCP) as an effective solution, as illustrated in Fig. \ref{introduction2}. The core idea is to mine co-occurring noise samples from unchanged image scenes and then enhance the model's robustness to such noise, thereby reducing false recognition in changed scenes. Scenes labeled as `unchanged' consist solely of unchanged pixels, making them a cleaner and more reliable source for identifying and addressing co-occurring noise. By applying adversarial prompting to these unchanged scenes, we systematically improve the model's ability to distinguish object changes from background variations.

Our AdvCP method consists of two phases: a) Adversarial Prompt Mining and b) Adversarial Sample Rectification. In a) Adversarial Prompt Mining, we introduce adversarial prompt perturbations using erroneous image-level classification labels. Specifically, we invert one-hot binary labels from `unchanged' to `changed', generating adversarially activated change localizations from originally unchanged images. These perturbations are applied online after each training iteration, exposing particular co-occurring sample pixels where background variations are likely misclassified as object changes.

Next, in b) Adversarial Sample Rectification, we incorporate adversarial pixel samples derived from these prompt-activated localizations into the training process. Specifically, an exponentially weighted moving average is used to construct an online global prototype from the current batch and all historical training data. This global prototype serves as an unbiased anchor, guiding the rectification of adversarial pixel samples and improving the model’s ability to distinguish between background variation noise and actual object changes.

The proposed AdvCP can be easily integrated into the current WSCD workflow, without introducing extra parameters. Experiments demonstrate that our AdvCP method significantly enhances six baselines, including ConvNet-based, Transformer-based, and SAM-based models. It achieves a maximum of 7.37\% IoU improvement on the WHU-CD dataset, 7.46\% on the LEVIR-CD dataset, and 6.56\% on the DSIFN-CD dataset. Furthermore, we demonstrate the applicability of AdvCP beyond WSCD, including fully-supervised change detection and other multi-class weakly-supervised dense prediction scenarios. AdvCP functions as a model-agnostic training strategy that can be applied to any network architecture without modifying the model itself. Our contributions can be summarized as follows.

\begin{itemize}
    \item We introduce a novel paradigm, AdvCP, which exposes the co-occurring pixel features of image-level classification models under weak supervision.
    
    \item We develop an online global prototype derived from all historical training data, acting as an unbiased anchor to guide the rectification of adversarial prompt samples during training.
    
    \item Our approach enhance the robustness of WSCD methods in a plug-and-play manner, achieving significant performance gains without introducing additional inference costs.
\end{itemize}

\section{Related Work}\label{sec:related}
\subsection{Weakly-Supervised Change Detection}
\par WSCD aims to identify specific pixel-level changes using only image-level annotations \cite{sakurada2020weakly}. Approaches such as \cite{Khan2017, ander2020} integrate convolutional and recurrent neural networks with conditional random fields, using change probability as a measure of confidence. FCD-GAN \cite{wu2023} employs a generative adversarial network combined with a fully convolutional network, similarly relying on the change probability to detect changes.

Recently, WSCD methods have utilized the family of CAMs for more convenient generation of change localization \cite{zhou2016learning, selvaraju2017grad}. For example, TransWCD \cite{transwcd} incorporates a label-gated constraint and a prior-based decoder to address discrepancies between predictions and weak labels, while MS-Former \cite{MSFormer} introduces a memory support mechanism to enhance WSCD. Additionally, CS-WSCDNet \cite{wang2023cs} leverages SAM \cite{kirillov2023segment} to create pixel-level pseudo labels, improving its applicability in various scenarios.

Although these works improve precision and recall, they still struggle with co-occurring background variations, often misclassifying lighting, weather, seasonal shifts, or human activities as object changes. These challenges are difficult to address under image-level supervision, where the model only knows whether a scene contains changes but lacks pixel-level guidance. In this work, we address this challenge cleverly by extracting co-occurring unchanged pixels from scenes labeled as `unchanged,' effectively enhancing the model’s robustness to co-occurring noise in predicting changed scenes.

\subsection{Pixel Rectification in Weakly-Supervised Dense Prediction}
\par WSCD can be regarded as segmenting pixel-level changes from paired images through change classification models. Therefore, it is often considered analogous to weakly-supervised semantic segmentation. 

In weakly-supervised semantic segmentation, rectifying pixel-level error predictions presents a significant challenge. OCR \cite{cheng2023out} utilizes the correlations of prior annotation and posterior prediction to tackle erroneous pixels that do not belong to the correct category. CPAL \cite{tang2024hunting} proposes context-aware prototypes that capture intra-class variations and align instance feature distributions with dense features, thereby tailoring the semantic attributes of diverse instances. These methods focus solely on correcting the inconsistency between pixel-level predictions and image-level labels, without directly addressing co-occurring background noise. 

W-OoD \cite{lee2022weakly} targets co-occurring background pixels, mistakenly identified as foreground. However, W-OoD depends on extra-labeled data, which is not always feasible or available in many real-world applications. In contrast, our AdvCP leverages the inherent characteristics of weakly-supervised tasks, and employs adversarial prompt perturbations to enhance model robustness. This process is executed without any increase in parameter overhead or annotation costs. 

\begin{figure*}
	\centering
	\includegraphics[scale=0.5
]{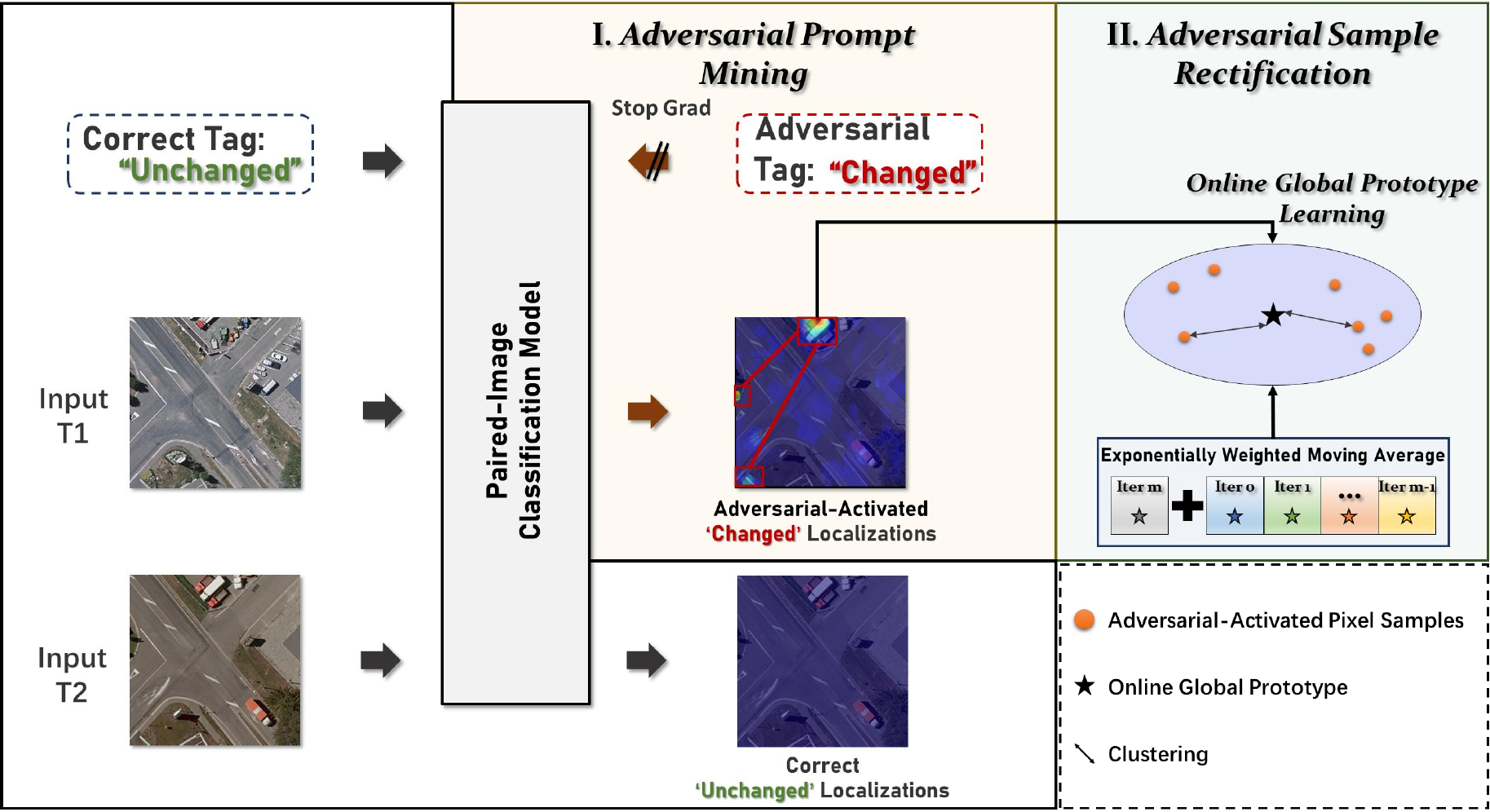}
	\caption{Framework of AdvCP. It includes two phases: a) Adversarial Prompt Mining: We use adversarial `unchanged' labels into `changed' to introduce adversarial prompt perturbations. b) Adversarial Sample Rectification: We employ the exponentially weighted moving average to construct an online global prototype from the current batch and all historical training data, which then serves as an unbiased anchor to guide the clustering of adversarial pixel samples.}
	\label{method1}
\end{figure*}
\subsection{Adversarial Sample Learning}

Adversarial samples present a significant challenge to the robustness of machine learning models \cite{szegedy2013intriguing}. A common strategy to mitigate this issue involves integrating adversarial learning into the training process, such as mining adversarial examples and incorporating them into the training dataset \cite{goodfellow2014explaining}. Madry \textit{et al.} \cite{madry2018towards} introduce multi-step projected gradient descent adversaries into training to strengthen model robustness. 

Recently, ACR \cite{Kweon_2023_CVPR} implements an adversarial concept between a classifier and an image reconstructor to improve weakly supervised semantic segmentation. Inspired by these works, we propose the AdvCP framework that leverages erroneous weak image labels to mine pixel-level class-ambiguous samples, addressing the issue of co-occurring background noise in WSCD.

\subsection{Prototype Learning for Dense Prediction}
Prototype learning has been studied in various computer vision and pattern recognition tasks \cite{he2019dynamic,snell2017prototypical,xu2020attribute} because of its ability to capture category characteristics in a compact feature representation. The key idea behind prototype learning is to summarize high-level information into class prototypes, often referred to as category feature centers. Typically, in dense prediction tasks, these prototypes are computed based on the current mini-batch \cite{zhou2022rethinking,wang2019panet,mao2022dpnet}, from an original frozen “once-extracted” global prototype \cite{xu2022semi,dong2018fewshot,shen2024adaptive}, or via memory-based approaches that store more representative samples over multiple batches \cite{zhao2022numberadaptive,liu2020partaware,yang2020mixture}. 

While these methods effectively leverage cluster centers to represent class-level features, they often struggle to balance stability, adaptability, and storage costs. In this work, we introduce an online global prototype that continually refines the prototype without storing all past data, leveraging an exponentially weighted moving average. Our prototype learning framework significantly reduces computational overhead and makes the approach more robust to adversarial samples.

\section{Methodology}\label{sec:method}
Our AdvCP consists of two parts: a) Adversarial Prompt Mining: the introduction of adversarial prompt perturbations is in Sec. \ref{sec:method1}, and b) Adversarial Sample Rectification: the integration of these adversarially activated pixel samples into the training process is in Sec. \ref{sec:method2}. An additional empirical analysis of the effectiveness of AdvCP is in Sec. \ref{sec:method3}. The framework of the proposed AdvCP is illustrated in Fig. \ref{method1}, and the algorithm is summarized in Algorithm \ref{alg:adversarial}.

We begin with the preliminary step of WSCD, describing how to derive pixel-level change predictions from a paired-image classification model.

\subsection{Preliminary}
Specifically, given a training batch for WSCD as \( B_{\text{train}} = \{(x_{t1}^n, x_{t2}^n, y_{cls}^n)\}_{n=1}^{N} \), including $N$ pairs of remote  sensing images $(x_{t1}, x_{t2})$, where the paired images \( x_{t1} \) and \( x_{t2} \) are captured at times \( t_1 \) and \( t_2 \) of the same scene, respectively. Their paired-image label $y_{cls} = \{\text{changed }(1), \text{unchanged }(0)\}$, indicates whether specific objects have changed or not between paired scene images.

The training batch data $ B_{\text{train}}$ is then fed into a change classification model with a two-stream input for the paired images $\{x_{t1}^n\}_{n=1}^N$ and $\{x_{t2}^n\}_{n=1}^N \in \mathbb{R}^{HW \times 3}$, and a single-stream output for change prediction maps \( \mathcal{P} \in \mathbb{R}^{N \times HW} \). Here, $HW$ represents the spatial size of the images. The change classification model is trained only with the constraint of a classification loss (\textit{e.g.}, cross-entropy loss).

Subsequently, pixel-level change predictions are derived from the change classification model. The last-layer feature maps $\mathcal{F} \in \mathbb{R}^{N \times HW \times D}$ are usually weighted by $w \in \mathbb{R}^{D \times 2}$ to generate class localization maps $\mathcal{C} \in \mathbb{R}^{N \times HW \times 2}$. Here, $w$ contains the two channel weights for the category of `changed' and `unchanged'. $D$ denotes the dimensions of the last-layer feature maps. Formally, the class localization maps $\mathcal{C}$ are generated as follows:
\begin{equation}
\label{cam}
 \mathcal{C} = \text{ReLU}(\sum_{j}^{D} \mathcal{F}^{(j)}w^{(j)}),
\end{equation}
where the $\text{ReLU}$ function eliminates the negative activation values. 

The weights $w$ of the feature maps are calculated either directly from the weight matrices $\mathcal{W}$ of the last classification layer, \textit{e.g.}, using CAM \cite{zhou2016learning}; or from the gradients $w=\frac{1}{HW} \sum  \frac{\partial \mathcal{F}^{(i)}} {\partial \hat{y}_{cls}}$ of the feature maps $\mathcal{F}$ with respect to the predicted classification score $\hat{y}_{cls} \in \mathbb{R}^N$, \textit{e.g.}, using Grad-CAM \cite{selvaraju2017grad}. Furthermore, the $\text{Max}$ normalization scales the values of the class localization maps to $[0,1]$, and the normalized class localization maps are $\mathcal{C}=\frac{\mathcal{C}^{(i)}}{\text{Max}(\mathcal{C}^{(i)})}$. The class localization maps \( \mathcal{C} \in \mathbb{R}^{N \times HW \times 2} \) consist of two category channels, \( \mathcal{C}_c \) and \( \mathcal{C}_{uc} \in \mathbb{R}^{N \times HW} \), representing the `changed' and `unchanged' categories, respectively.

Finally, the binary pixel-level change predictions $\mathcal{P} \in \mathbb{R}^{N \times HW}$ are calculated as follows: 
\begin{equation}
\label{threshold}
\mathcal{P}^{(i)}= \begin{cases}
\text{changed} \; (1), & \textit{if} \; \mathcal{C}_c^{(i)} \geq  \mathcal{C}_{uc}^{(i)}\\
\text{unchanged} \; (0), & \textit{otherwise}
\end{cases}. 
\end{equation}
The above WSCD workflow exploits the classification models' tendency to focus on regions that are most significant to make a classification decision of `changed' \cite{zhou2016learning}. However, models based on paired-image classification struggle to ensure the accuracy and robustness of pixel-level predictions. Co-occurring background variations are often misidentified as object changes and cannot be effectively distinguished under image-level weak supervision.

\subsection{Adversarial Prompt Mining} \label{sec:method1}
\begin{figure*}[!ht]
	\centering
\includegraphics[width=0.90\linewidth]{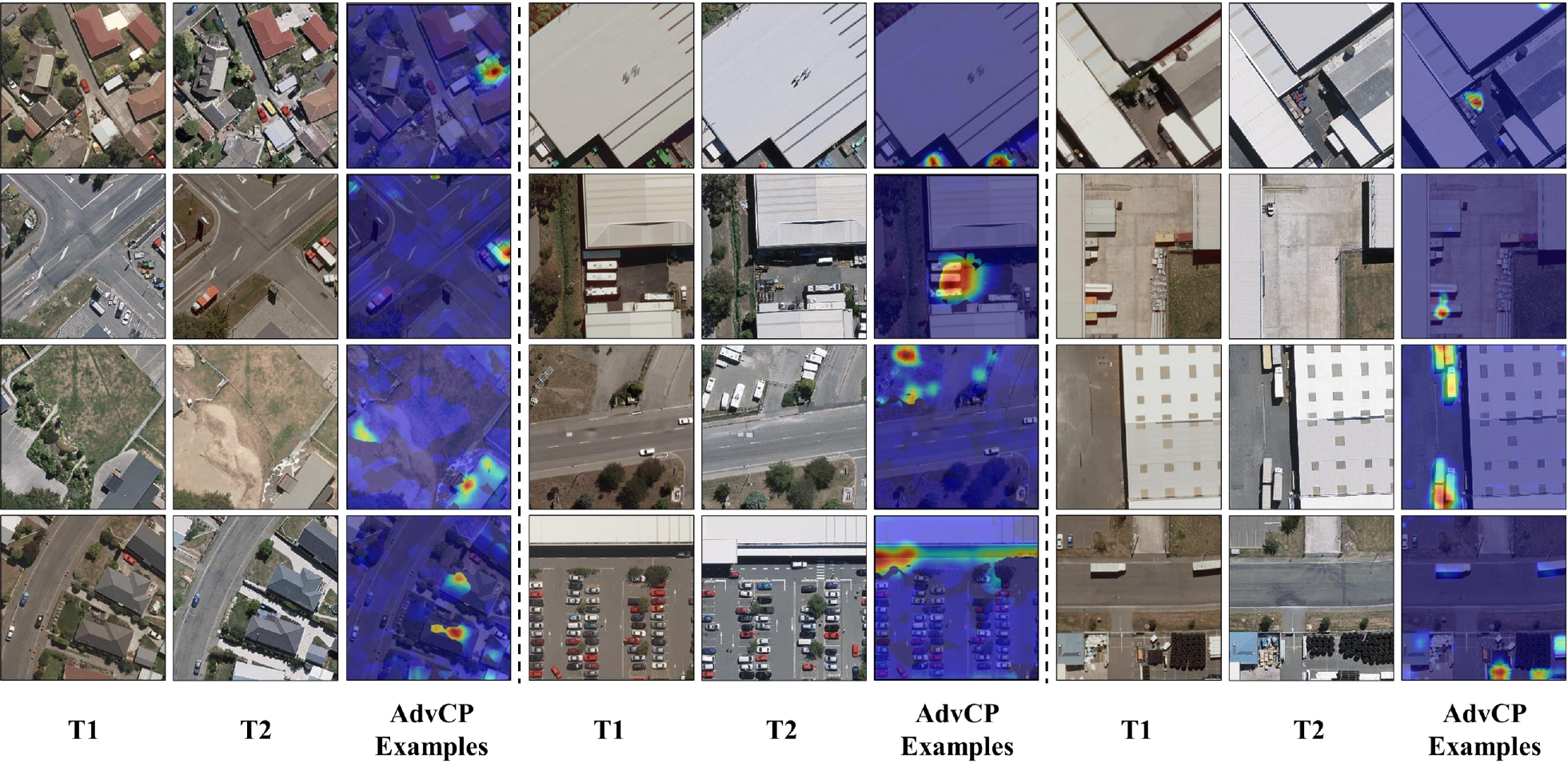}
	\caption{Examples of adversarial prompt perturbations. The color intensity in attention heatmaps reflects the important of features as considered by the models. Our AdvCP exposes indistinguishable background variations from `unchanged' scene images, with the adversarial `changed' classification labels. }
	\label{method2}
\end{figure*}

In this section, we tailor adversarial prompt perturbations to expose these pixel-level indistinguishable background variations under image-level weak supervision. After each training iteration, we apply the adversarial `changed' label to prompt the classification model to produce change discriminative localizations in paired-image examples labeled `unchanged'. 

For efficient processing within the current WSCD pipeline, we generate change localization maps \( \mathcal{C}_{c} \) for all paired-image examples. Then, we perform an XOR operation between the all-change localization maps \( \mathcal{C}_{c} \) and the original localization maps \( \mathcal{C} \). This XOR operation effectively extracts adversarially activated pixel samples, \textit{i.e.,} background variations. Examples of adversarial prompt perturbations are shown in Fig.~\ref{method2}.

Specifically, we first assign the `changed' classification label $y_{cls}^n = 1$ to all $N$ paired images in the training batch $ B_{\text{train}}$ after each training step. This label reassignment prompts the change classification model to use the change weights $w_{c} \in \mathbb{R}^{D}$ to calculate the corresponding all-change localization maps \(\mathcal{C}_{c}\in \mathbb{R}^{N \times HW}\), analogous to Eq. \ref{cam}:
\begin{equation}
\label{c_c}
\mathcal{C}_{c} = \text{ReLU}(\sum_{i}^{D} \mathcal{F}^{(i)}w_{c}^{(i)}).
\end{equation}
The change weight matrices \(w_c\) correspond to the weights $\mathcal{W}_{c}$ with respect to the `changed' category. Alternatively, these weights can be derived from the gradients by $w_c = \frac{1}{HW} \sum \frac{\partial \mathcal{F}^{(i)}}{\partial \hat{y}_{cls}}$, where \(\hat{y}_{cls} \in \mathbb{R}^N\) is the `changed' classification score.

Next, we derive the mask $\mathcal{M}_{adv} \in \mathbb{R}^{N \times HW}$ of adversarially activated pixel samples by performing an XOR operation between the all-change localization maps $\mathcal{C}_{c}$ and the original localization maps $\mathcal{C}$: \begin{equation}
\mathcal{M}_{adv} = \mathcal{C}_{c} \oplus \mathcal{C}.
\end{equation}
Here,  \(\mathcal{C}_{c} \in \mathbb{R}^{N \times HW}\) is conducted with the operation of channel duplication into  \(\mathcal{C}_{c} \in \mathbb{R}^{N \times HW \times 2}\). 

Then, we extract the pixel-level features $\mathcal{F}_{adv}$ of adversarial-activated pixel samples, from the feature maps $\mathcal{F}$ by element-wise multiplying: 
\begin{equation}
\mathcal{F}_{adv} =\mathcal{F} \odot \mathcal{M}_{adv}.
\end{equation}
In practice, \( \mathcal{F} \) may go through an extension of linear interpolation, ensuring its size consistency with $\mathcal{M}_{adv}$.

\begin{algorithm}
\small{
\caption{Adversarial Class Prompting}\label{alg:adversarial}
\KwIn{
    Training batch $B_{\text{train}} = \{(x_{t1}^n, x_{t2}^n, y_{cls}^n)\}_{n=1}^N$, 
    feature maps $\mathcal{F}$, localization maps $\mathcal{C}$, 
    predictions $\mathcal{P}$, historical prototype $p_{uc}^{(m-1)}$, 
    momentum $\lambda$, loss weight $\alpha$
}
\KwOut{
    Updated prototype $p_{uc}^{(m)}$, adversarial loss $\mathcal{L}_{adv}$
}

\textbf{Initialize:} 
$p_{uc}^{(0)} \gets \mathbf{0}$

\ForEach{training step $m$}{
    
    \textbf{Step 1: Generate adversarial prompt perturbations}\;
    \For{$n \gets 1$ to $N$}{
        $y_{cls}^n \gets 1$ \tcp*{Assign `changed' label}
    }
    
    $\mathcal{C}_c \gets \text{ReLU}\left(\sum_{j=1}^D \mathcal{F}^{(j)} w_c^{(j)}\right)$
    
    $\mathcal{M}_{adv} \gets \mathcal{C}_c \oplus \mathcal{C}$\tcp*{Generate adversarial mask}

    \textbf{Step 2: Extract Adversarial Features}\;
    $\mathcal{F}_{adv} \gets \mathcal{F} \odot \mathcal{M}_{adv}$

    \textbf{Step 3: Compute Batch-Wise Unchanged Prototype}\;
    $\mathcal{F}_{uc}^{(m)} \gets \mathcal{F}^{(m)} \odot (1 - \mathcal{P}^{(m)})$ 
    $N_{uc} \gets \text{CountNonZero}(1 - \mathcal{P}^{(m)})$
    \eIf{$N_{uc} > 0$}{
        $f_{uc}^{(m)} \gets \text{Mean}(\mathcal{F}_{uc}^{(m)})$
    }{
        $f_{uc}^{(m)} \gets \mathbf{0}$
    }

    \textbf{Step 4: Update Online Global Prototype}\;
    $p_{uc}^{(m)} \gets (1 - \lambda) p_{uc}^{(m-1)} + \lambda f_{uc}^{(m)}$

    \textbf{Step 5: Compute AdvCP Loss}\;
    $\mathcal{L}_{adv} \gets \|\mathcal{F}_{adv} - p_{uc}^{(m)}\|_2^2$
}

\textbf{Output:} $p_{uc}^{(m)}$, $\mathcal{L}_{adv}$
}
\end{algorithm}

\subsection{Adversarial Sample Rectification} \label{sec:method2}
\begin{figure}
	\centering
\includegraphics[width=0.85\linewidth]{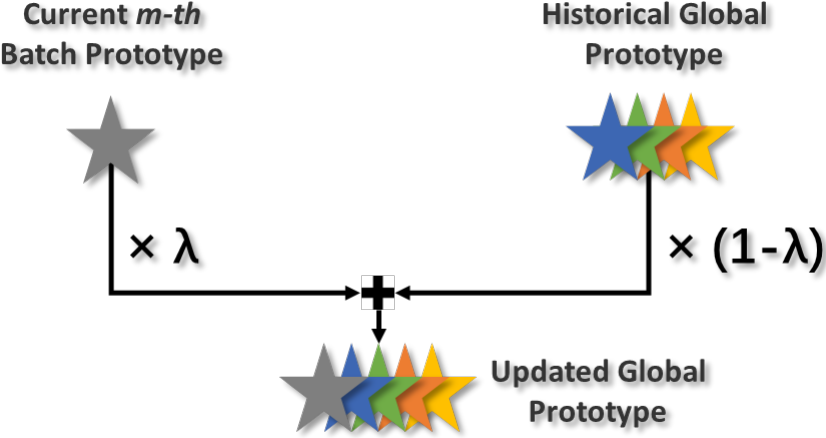}
	\caption{Calculation summary of the online global prototype. The online global unchanged prototype is computed using an exponentially weighted moving average that integrates the current \( m \)-th batch with all historical \( m-1 \) batches with a momentum coefficient \( \lambda \).}
	\label{method3}
\end{figure}

In this section, we further incorporate these adversarial-activated pixels into training and improve the robustness of WSCD under image-level weak supervision. In a non-parametric approach, we construct an online global prototype from all historical and current training data using an exponentially weighted moving average, as summarized in Fig. \ref{method3}. This online global prototype serves as an unbiased anchor for adversarial-activated pixel samples.

Specifically, let $\mathcal{F}_{uc}^{(m)}$ denote pixel-level feature maps of background regions, in the current $m$-th iteration. We calculate $\mathcal{F}_{uc}^{(m)}$ by element-wise multiplying it with the binary pixel-level change predictions $\mathcal{P}^{(m)}$: 
\begin{equation}
\mathcal{F}_{uc}^{(m)} =\mathcal{F}^{(m)} \odot (1-\mathcal{P}^{(m)}).
\end{equation}
$\mathcal{P}^{(m)}$ is calculated in Eq. \ref{threshold}. Then, we obtain the unchanged prototype $f_{uc}^{(m)} \in \mathbb{R}^{D}$ by averaging the pixel-level feature vectors within the unchanged feature maps: $f_{uc}^{(m)} = \frac{1}{N_{uc}} \sum \mathcal{F}_{uc}^{(m)}$. 

Due to the randomness of sample selection in a batch, the batch-wise unchanged prototype $f_{uc}^{(m)}$ may fail to represent the center of the unchanged category accurately. \textcolor{black}{This issue becomes more serious in scenarios with highly variable distributions, where the sample diversity across batches can lead to instability, preventing the model from capturing reliable clustering centers.} Thus, we further update the batch-wise unchanged prototype $f_{uc}^{(m)}$ to the online global unchanged prototype $p_{uc}^{(m)}$, via the exponentially weighted moving average:
\begin{equation} 
\label{ctr}
p_{uc}^{(m)} = (1-\lambda) \cdot p_{uc}^{(m-1)}  + \lambda \cdot f_{uc}^{(m)}.
\end{equation}
The online global unchanged prototype $p_{uc}^{(m)}$ is calculated by combining the current $m$-th batch with all the previous $m-1$ batches. Here, $\lambda$ is a momentum coefficient used to control the update rate, balancing the contributions of the historical training data and the current batch data.

Subsequently, we guide the feature mappings \( \mathcal{F}^{(m)}_{adv}\) of adversarial-activated pixels, to cluster into the online global unchanged prototype $p_{uc}^{(m)}$. Mathematically, it is formulated as the AdvCP loss $\mathcal{L}_{adv}$, constrained by the L2 distance:
\begin{equation} 
\mathcal{L}^{(m)}_{adv} =  \| \mathcal{F}^{(m)}_{adv} - p_{uc}^{(m)} \|_2^2.
\end{equation}
Here, the online global unchanged prototype $p_{uc}^{(m)}$ is broadcast into the size of \( \mathcal{F}^{(m)}_{adv}\). $\mathcal{L}_{adv}$ draws adversarial-activated pixels (\textit{i.e.}, features of indistinguishable background variations) closer to the unchanged class center, thus establishing a discriminative margin between changed and unchanged pixel features in the embedding space.

The overall loss $\mathcal{L}$, which consists of the binary cross-entropy classification loss $\mathcal{L}_{cls}$ and the AdvCP loss $\mathcal{L}_{adv}$, is given by:
\begin{equation} 
\label{total_loss}
\mathcal{L} = \mathcal{L}_{cls} + \alpha \mathcal{L}_{adv},
\end{equation}
where $\alpha > 0$ is an impact factor balancing the contribution of $\mathcal{L}_{adv}$. \textcolor{black}{In practice, the AdvCP constraint is delayed and introduced after the classification constraint. This allows the model to first learn stable features and remove spurious noise before applying adversarial perturbations.}

\subsection{Empirical Insights into Effectiveness in the Latent Space}
\label{sec:method3}
\begin{figure}[ht!]
    \centering
    \includegraphics[scale=0.73]{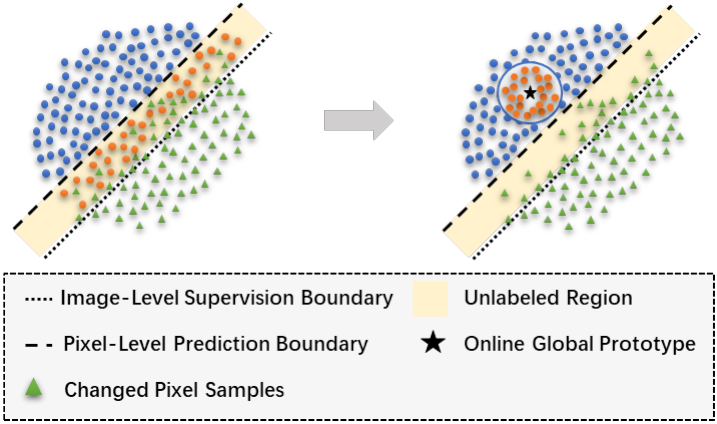}
    \caption{Effectiveness analysis of our AdvCP method in the latent space. The challenge arises from background variation noise, which can be seen as hard samples from the unchanged category. These hard samples are often misclassified as changed due to the limited guidance from image-level supervision. Our AdvCP method helps to separate these hard pixel samples from the prediction boundary, leading to more accurate predictions under weak supervision.}
    \label{appendix5}
\end{figure}

In this section, we provide an empirical analysis of the effectiveness of the proposed AdvCP method within the latent space. As illustrated in Fig. \ref{appendix5}, we simplify other unrelated factors to concentrate specifically on the issue of co-occurring background noise. 

Under weak supervision, image-level classification labels tend to highlight only pixels with high category confidence.
This background noise can be considered as hard samples belonging to the unchanged category, which are more class-ambiguous because the image-level supervision does not provide enough pixel-level guidance.

As a result, co-occurring background noise pixels remain unsupervised within the regions between the supervision boundary and the prediction boundary, as shown in Fig. \ref{appendix5}. Without explicit guidance, the model may confuse unchanged pixel noise with changed samples, leading to inaccurate predictions.

Our AdvCP method addresses this issue by first using adversarial prompting to explicitly identify these overlooked noise samples. Once these hard samples are located, the method utilizes an online global prototype to push them away from the decision boundary. This approach helps establish a clearer separation between the changed and unchanged categories, thereby enhancing prediction accuracy under weak supervision. 

\section{Experiments}
\begin{table*}[!ht]
 \setlength{\tabcolsep}{5pt}
\renewcommand{\arraystretch}{1.2}
\centering
\caption{Quantitative improvements over baselines and comparisons with weakly-supervised change detection methods. F1 score (\%), OA (\%), IoU (\%) are reported.}
\label{WSCD}\small
\begin{tabular}{clccclccclcccl} 
\hline\hline
\multirow{2}{*}{\textbf{Method}} &  & \multicolumn{3}{c}{\textbf{WHU-CD}} & \multicolumn{1}{c}{} & \multicolumn{3}{c}{\textbf{LEVIR-CD}} &  & \multicolumn{3}{c}{\textbf{DSIFN-CD}} & \multicolumn{1}{c}{} \\ 
\cline{2-14}
 &  & \textbf{F1} & \textbf{OA} & \textbf{IoU} &  & \textbf{F1} & \textbf{OA} & \textbf{IoU} &  & \textbf{F1} & \textbf{OA} & \textbf{IoU} &  \\ 
\hline
FCD-GAN          & & 56.45 & 78.98 & 39.32 & & 43.08 & 94.49 & 30.45 & & 40.26 & 73.27 & 29.94 & \\
BGMix            & & 62.40 & 84.40 & 42.70 & & 51.98 & 93.52 & 38.12 & & 45.80 & 77.96 & 31.94 & \\
MS-Former        & & 65.03 & 94.40 & 50.04 & & 61.24 & 94.72 & 43.07 & & 52.55 & 83.32 & 37.04 & \\ 
\hline
WCDNet           & & 42.33 & 86.26 & 24.31 & & 32.56 & 88.45 & 18.79 & & 29.77 & 72.59 & 17.26 & \\
\rowcolor[rgb]{0.925,0.925,0.925} \textbf{+AdvCP (Ours)} 
 & & \textbf{49.12\tiny\textcolor{red}{+6.79}} & \textbf{93.00\tiny\textcolor{red}{+6.74}} & \textbf{31.68\tiny\textcolor{red}{+7.37}} 
 & & \textbf{35.66\tiny\textcolor{red}{+3.10}} & \textbf{93.82\tiny\textcolor{red}{+5.37}} & \textbf{24.80\tiny\textcolor{red}{+6.01}}
 & & \textbf{35.47\tiny\textcolor{red}{+5.70}} & \textbf{75.20\tiny\textcolor{red}{+2.61}} & \textbf{23.13\tiny\textcolor{red}{+5.87}} & \\
\hline
SEAM             & & 44.45 & 84.17 & 27.14 & & 35.81 & 89.03 & 27.07 & & 27.56 & 71.42 & 20.01 & \\
\rowcolor[rgb]{0.925,0.925,0.925} \textbf{+AdvCP (Ours)} 
 & & \textbf{50.20\tiny\textcolor{red}{+5.75}} & \textbf{91.41\tiny\textcolor{red}{+7.24}} & \textbf{31.90\tiny\textcolor{red}{+4.76}}
 & & \textbf{39.74\tiny\textcolor{red}{+3.93}} & \textbf{94.46\tiny\textcolor{red}{+5.43}} & \textbf{31.41\tiny\textcolor{red}{+4.34}}
 & & \textbf{33.50\tiny\textcolor{red}{+5.94}} & \textbf{76.10\tiny\textcolor{red}{+4.68}} & \textbf{25.13\tiny\textcolor{red}{+5.12}} & \\
\hline
MCTformer         & & 60.64 & 95.02 & 51.45 & & 55.17 & 93.05 & 39.54 & & 49.33 & 80.21 & 32.69 & \\
\rowcolor[rgb]{0.925,0.925,0.925} \textbf{+AdvCP (Ours)} 
 & & \textbf{68.34\tiny\textcolor{red}{+7.70}} & \textbf{97.22\tiny\textcolor{red}{+2.20}} & \textbf{56.73\tiny\textcolor{red}{+5.28}} 
 & & \textbf{60.12\tiny\textcolor{red}{+4.95}} & \textbf{96.34\tiny\textcolor{red}{+3.29}} & \textbf{43.51\tiny\textcolor{red}{+3.97}}
 & & \textbf{55.78\tiny\textcolor{red}{+6.45}} & \textbf{82.36\tiny\textcolor{red}{+2.15}} & \textbf{39.25\tiny\textcolor{red}{+6.56}} & \\
\hline
TransWCD         & & 68.73 & 97.17 & 52.36 & & 60.08 & 95.56 & 42.94 & & 53.41 & 83.05 & 36.44 & \\
\rowcolor[rgb]{0.925,0.925,0.925} \textbf{+AdvCP (Ours)} 
 & & \textbf{74.45\tiny\textcolor{red}{+5.72}} & \textbf{98.35\tiny\textcolor{red}{+1.18}} & \textbf{59.30\tiny\textcolor{red}{+6.94}} 
 & & \textbf{66.03\tiny\textcolor{red}{+5.95}} & \textbf{97.01\tiny\textcolor{red}{+1.45}} & \textbf{50.40\tiny\textcolor{red}{+7.46}}
 & & \textbf{58.67\tiny\textcolor{red}{+5.26}} & \textbf{84.45\tiny\textcolor{red}{+1.40}} & \textbf{42.14\tiny\textcolor{red}{+5.70}} & \\
\hline
\rowcolor{white} CS-WSCDNet& & 72.84 & 97.91 & 57.30 & & 66.83 & 98.39 & 48.27 & & 56.90 & 84.95 & 40.07 & \\
\rowcolor[rgb]{0.925,0.925,0.925} \textbf{+AdvCP (Ours)} 
 & & \textbf{76.34\tiny\textcolor{red}{+3.50}} & \textbf{98.26\tiny\textcolor{red}{+0.35}}& \textbf{61.48\tiny\textcolor{red}{+4.18}} 
 & & \textbf{69.04\tiny\textcolor{red}{+2.21}} & \textbf{98.86\tiny\textcolor{red}{+0.47}} & \textbf{49.56\tiny\textcolor{red}{+1.29}} 
 & & \textbf{58.81\tiny\textcolor{red}{+1.91}} & \textbf{85.54\tiny\textcolor{red}{+0.59}} & \textbf{42.23\tiny\textcolor{red}{+2.16}} & \\
\hline
\rowcolor{white} 
S2C& & 73.73 & 98.02 & 58.40 & & 67.94 & 98.49 & 48.18 & & 56.41 & 85.36 & 39.58 & \\
\rowcolor[rgb]{0.925,0.925,0.925} \textbf{+AdvCP (Ours)} 
 & & \textbf{77.25\tiny\textcolor{red}{+3.52}} & \textbf{98.51\tiny\textcolor{red}{+0.49}} & \textbf{62.39\tiny\textcolor{red}{+3.99}} 
& & \textbf{70.06\tiny\textcolor{red}{+2.12}}& \textbf{98.95\tiny\textcolor{red}{+0.46}}& \textbf{50.45\tiny\textcolor{red}{+2.27}} 
& & \textbf{58.90\tiny\textcolor{red}{+2.49}}& \textbf{85.75\tiny\textcolor{red}{+0.39}} & \textbf{43.10\tiny\textcolor{red}{+3.52}}& \\
\hline\hline
\end{tabular}
\end{table*}
\begin{figure*}
	\centering
	\includegraphics[width=\linewidth]{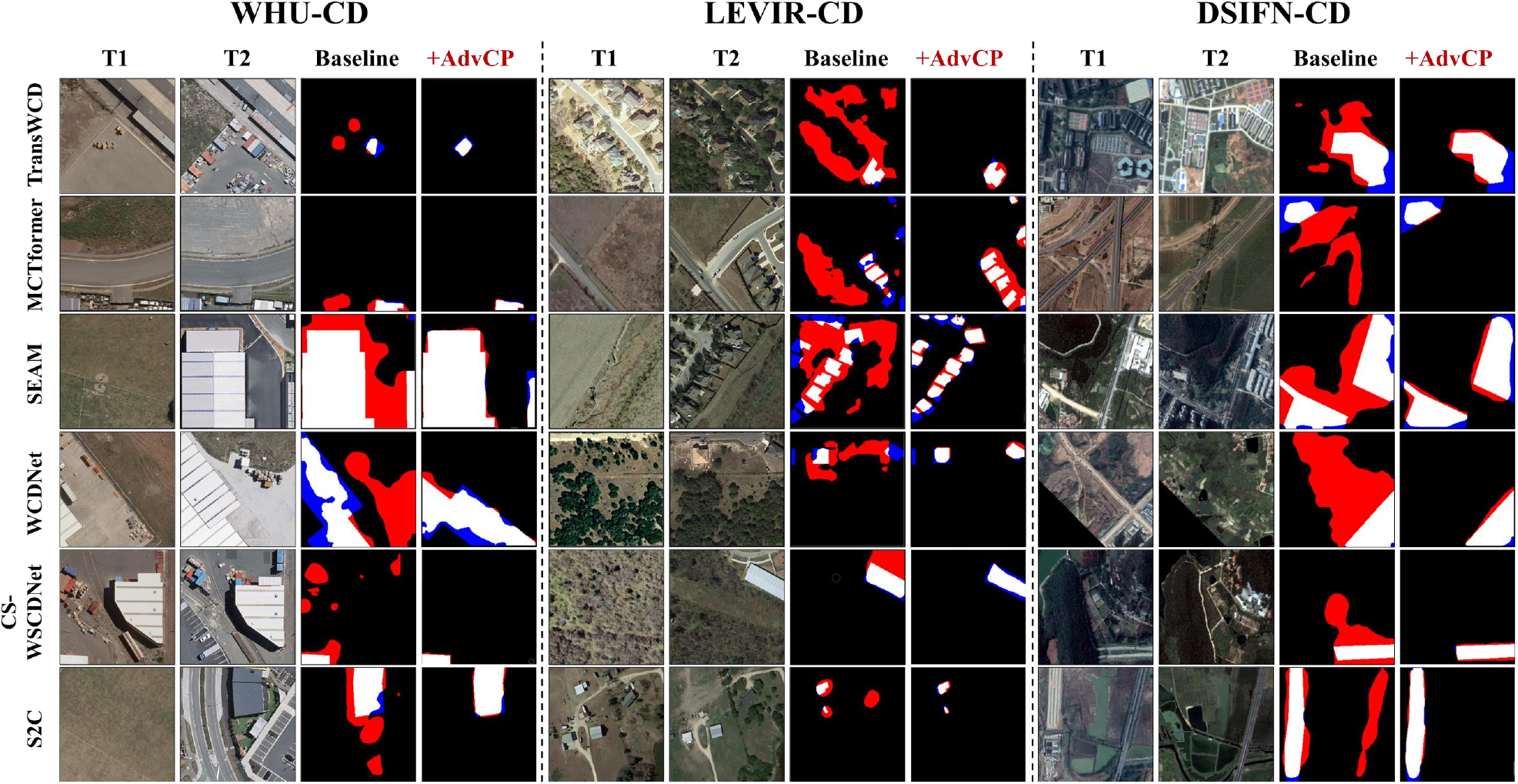}
	\caption{Qualitative improvements of our AdvCP on six baseline methods. For clarity, in the predictions, \textcolor{red}{false positive (erroneous changed) pixels} are marked in red color, and \textcolor{blue}{false negative (erroneous unchanged) pixels} are marked in  color. Our AdvCP clearly enhances the baseline methods, although the results are also influenced by their inherent limitations.}
	\label{experiment0}
\end{figure*}

\subsection{Experimental Setup}
\noindent\textbf{Datasets.} We perform our experiments on three widely used CD datasets: the LEVIR-CD dataset \cite{chen2020spatial}, the WHU-CD \cite{ji2018fully}, and the DSIFN-CD dataset \cite{zhang2020deeply}. We follow the same data partitioning used in previous studies \cite{Huang2023} to ensure a fair comparison.

The LEVIR-CD dataset is a large-scale building change detection (CD) dataset, consisting of high-resolution (0.5 m) remote sensing image pairs, collected between 2012 and 2018. The dataset comprises a total of 38,639 changed instances, with each image having a size of 256×256 pixels. The data is divided into 70\% for training, 10\% for validation, and 20\% for testing. 

The WHU-CD dataset focuses on urban change detection and contains a pair of very high-resolution (0.3 m) aerial images, each measuring 32,507×15,354 pixels, taken in 2012 and 2016. The images are divided into 256×256 pixel segments, with the data split into 60\% for training, 20\% for validation, and 20\% for testing.

\textcolor{black}{The DSIFN-CD dataset is a cross-domain benchmark that broadens the scope of change detection to include a variety of land-cover objects. It contains six pairs of high-resolution (2m) satellite images from six different cities. Following the default 256×256 pixel segmentation, the dataset includes 32,908 changed instances. We follow the original dataset partitioning for cross-domain change detection evaluation, as defined in the official release, comprising 4,400 training samples, 1,360 validation samples, and 192 test samples. The training and validation samples are drawn from the same source regions, while the test samples are collected from a different geographic area.}

\noindent\textbf{Evaluation Protocol.} To be consistent with previous works \cite{Huang2023}, we use three commonly recognized evaluation metrics: overall accuracy (OA), intersection over union (IoU), and F1 score. The F1 score, which combines precision and recall, is the primary indicator for change detection \cite{Zhang2022}. 

\noindent\textbf{Baselines and Implementation Details.} AdvCP is evaluated on six baseline methods: ConvNet-based WCDNet \cite{ander2020} and SEAM \cite{wang2020self}, Transformer-based MCTformer \cite{xu2022multi} and TransWCD \cite{transwcd}, and SAM-based CS-WSCDNet \cite{wang2023cs} and S2C \cite{kweon2024sam}. WCDNet, TransWCD, and CS-WSCDNet are weakly-supervised change detection methods. SEAM, MCTformer, and S2C are originally designed for weakly-supervised semantic segmentation.

For TransWCD and CS-WSCDNet, we evaluate our AdvCP directly without further modifications. For WCDNet, we have reproduced it with CAMs, leading to an improvement in performance compared to the original version. \textcolor{black}{For SEAM, MCTformer, and S2C, which perform well on semantic segmentation tasks, we modify the single-image input of these segmentation methods to accept double-image inputs by adding a 1x1 convolutional layer (without ReLU) before the original backbone, making them suitable for the task of weakly-supervised change detection while still using their original pre-trained weights. This modification makes the model transition from the semantic segmentation task to the change detection task, capturing changes in remote sensing.}

\textcolor{black}{We follow the default training and inference procedures of the baselines, including data augmentation, pre-trained weights, learning rate decay policy, weight decay, and optimizer settings, with a batch size of 16 applied to all three datasets. The training schedule varies by dataset and model: for weakly-supervised change detection baselines, the number of iterations is 8,000, 12,000, and 20,000 for WHU-CD, DSIFN-CD, and LEVIR-CD, respectively; for the weakly-supervised semantic segmentation baselines, the modified SEAM is trained for 30 epochs, and MCTformer and S2C for 50 epochs.}  

\textcolor{black}{Our AdvCP is incorporated into the back-propagation process after 200 initial iterations, with the optimal impact factor \( \alpha \) of \( \mathcal{L}_{adv} \) in Eq. \ref{total_loss} set to 1.0, 0.5, and 0.5 for WHU-CD, LEVIR-CD, and DSIFN-CD, respectively.}

\begin{figure}
	\centering
	\includegraphics[width=0.955\linewidth]{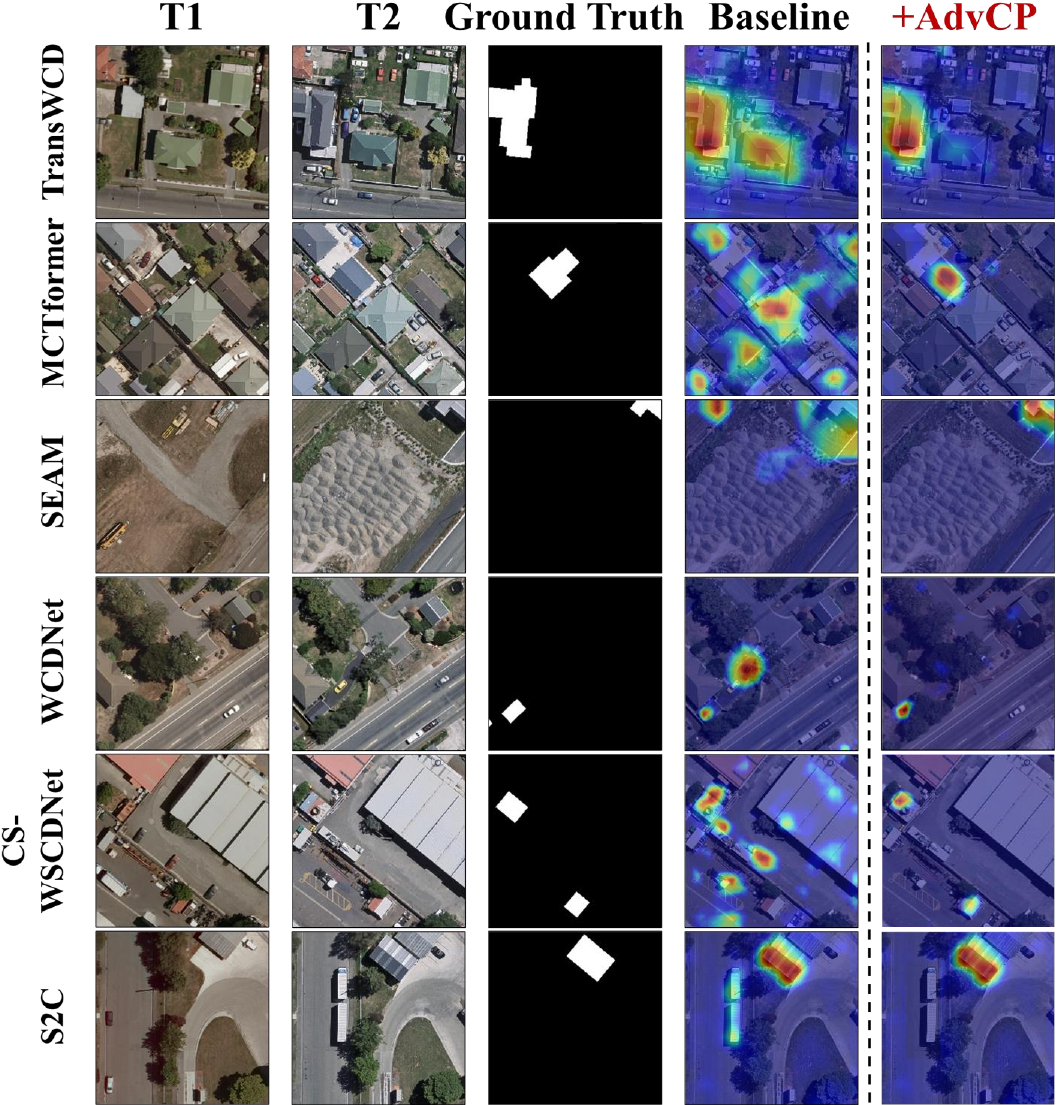}
	\caption{Heatmap comparisons of our AdvCP with six baseline methods on the \textbf{WHU-CD} dataset. The color intensity in the heatmaps reflects the saliency features as determined by the models.}
	\label{experiment1_1}
\end{figure}
\begin{figure}
	\centering
	\includegraphics[width=0.955\linewidth]{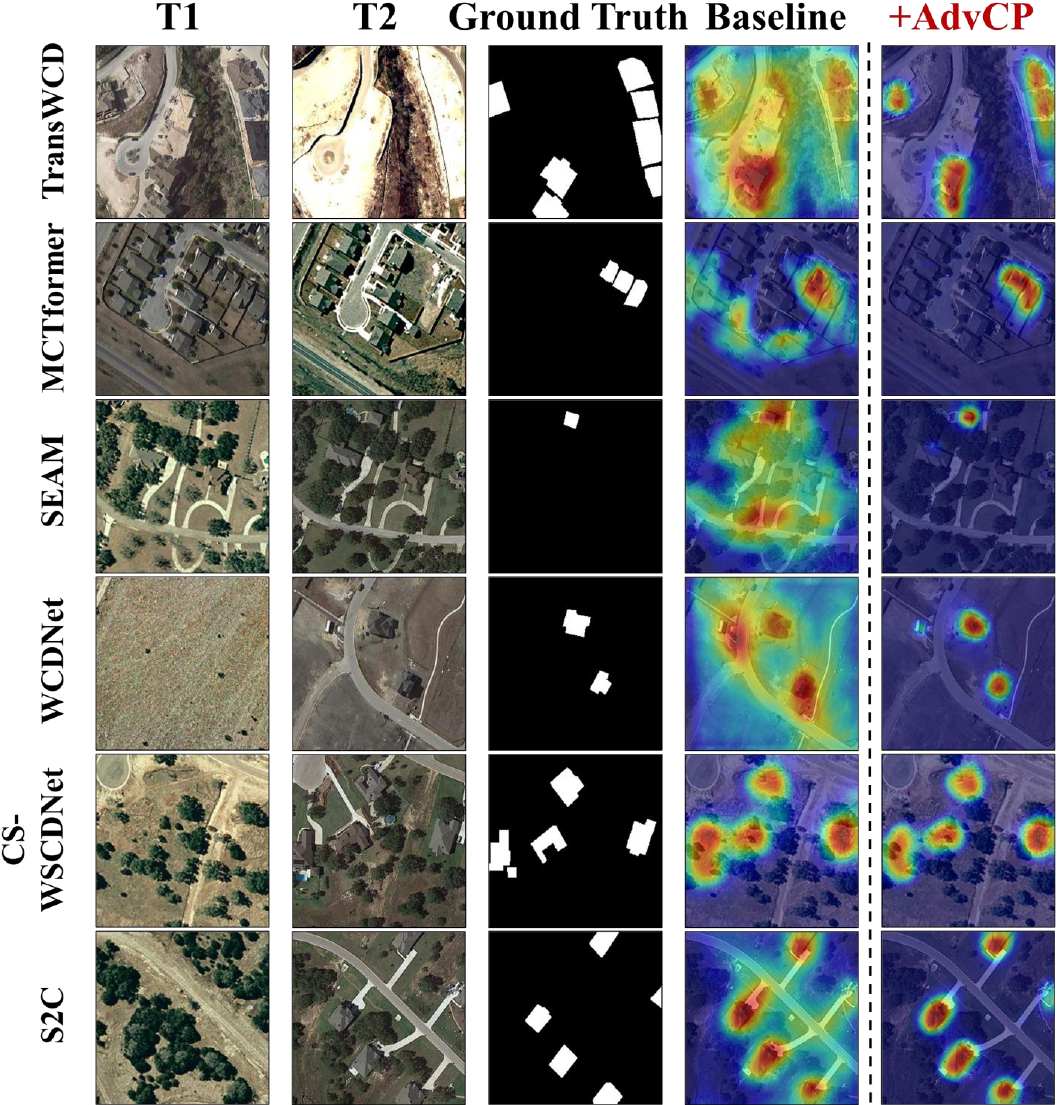}
	\caption{Heatmap comparisons of our AdvCP with six baseline methods on the \textbf{LEVIR-CD} dataset. The color intensity in the heatmaps reflects the saliency features as determined by the models. }
	\label{experiment1_2}
\end{figure}
\subsection{Comparison to State-of-the-Art Methods}
To verify the effectiveness of WSCD, we evaluate our AdvCP on top of six baselines: WCDNet \cite{ander2020}, SEAM \cite{wang2020self}, MCTformer \cite{xu2022multi}, TransWCD \cite{transwcd}, CS-WSCDNet \cite{wang2023cs}, and S2C \cite{kweon2024sam}. We also compare the quantitative results with several existing WSCD methods:  FCD-GAN \cite{wu2023}, BGMix \cite{Huang2023}, and the approach by Kalita \textit{et al.} \cite{kal2021}.

\noindent\textbf{Quantitative Comparison}. As shown in Table \ref{WSCD}, our AdvCP outperforms existing state-of-the-art methods by a significant margin. Remarkably, when combined with the SAM-based S2C baseline, our AdvCP sets new benchmarks on the WHU-CD dataset with 77.25\% F1 score, 98.51\% OA, and 62.39\% IoU; on the LEVIR-CD dataset with 70.06\% F1 score, 98.95\% OA, and 50.45\% IoU; and on the DSIFN-CD dataset with 58. 90\% F1 score, 85. 75\% OA, and 43. 10\% IoU.

Specifically, applying AdvCP to S2C results in notable improvements: +3.52\% in F1 score, +0.49\% in OA, and +3.99\% in IoU on WHU-CD; +2.12\% in F1 score, +0.46\% in OA, and +2.27\% in IoU on LEVIR-CD; and +2.49\% in F1 score, +0.39\% in OA, and +3.52\% in IoU on DSIFN-CD. These gains demonstrate the substantial enhancement brought by AdvCP, even for improving the performance of foundational models in WSCD.

For TransWCD, the impact of AdvCP is even more pronounced, with increases of +5.72\% in F1 score, +1.18\% in OA, and +6.94\% in IoU on WHU-CD; +5.95\% in F1 score, +1.45\% in OA, and +7.46\% in IoU on LEVIR-CD; and +5.26\% in F1 score, +1.40\% in OA, and +5.70\% in IoU on DSIFN-CD.

These results highlight AdvCP's versatility across different neural network architectures, with its performance on the more challenging DSIFN-CD dataset further demonstrating its effectiveness. In particular, AdvCP achieves significant gains in F1 score across six different baselines on the DSIFN-CD dataset, with improvements of +5.70\%, +5.94\%, +6.45\%, +5.26\%,  +1.91\%, and +2.49\%. These enhancements demonstrate a substantial improvement in distinguishing between changed and unchanged pixels, especially in more complex scenarios. 

\begin{figure}
	\centering
	\includegraphics[width=0.955\linewidth]{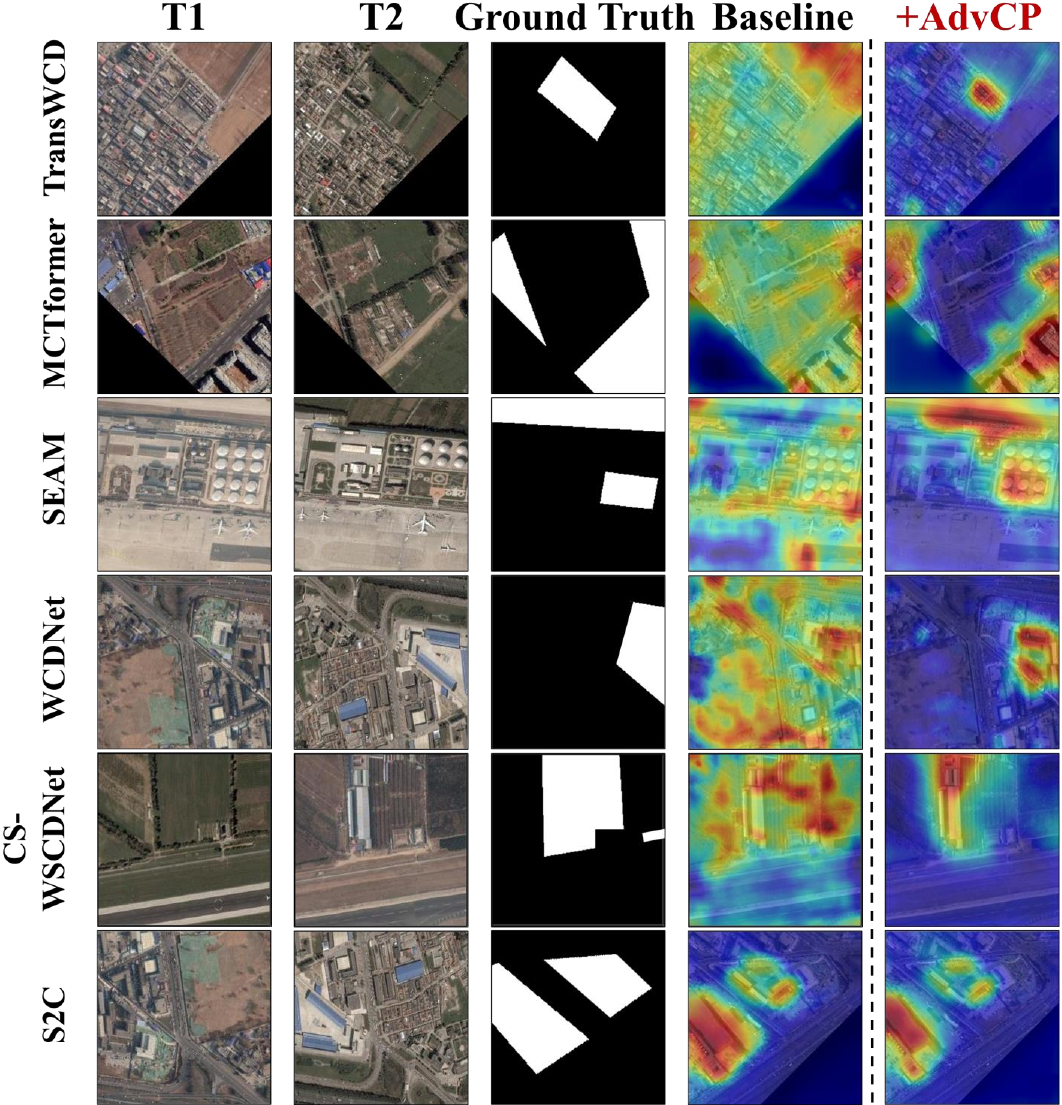}
	\caption{Heatmap comparisons of our AdvCP with six baseline methods on the \textbf{DSIFN-CD} dataset. The color intensity in the heatmaps reflects the saliency features as determined by the models. }
	\label{experiment1_3}
\end{figure}

\noindent\textbf{Qualitative Comparison.} We demonstrate the qualitative improvements achieved by applying AdvCP to six baseline methods. The prediction results are shown in Fig. \ref{experiment0}, with heatmap comparisons presented as follows: Fig. \ref{experiment1_1} for the WHU-CD dataset, Fig. \ref{experiment1_2} for the LEVIR-CD dataset, and Fig. \ref{experiment1_3} for the DSIFN-CD dataset. Our AdvCP significantly reduces background variation noise in pixel-level change localization, highlighting its ability to boost robustness and generalization.

Besides addressing the problem of co-occurring background variations, our AdvCP method significantly minimizes false positives and increases true positive detections for WSCD. These improvements include refining the boundaries of changed objects, ensuring more accurate delineation. For instance, as shown in the last row of Fig. \ref{experiment1_2} for the LEVIR-CD dataset, and the second-to-last row of Fig. \ref{experiment1_3} for the DSIFN-CD dataset, our method effectively enhances detection accuracy. These advancements are crucial for applications that demand high accuracy in change detection, such as urban planning and environmental monitoring.

\subsection{Feature Distribution Visualization}
\begin{figure*}[!ht]
	\centering
	\includegraphics[width=0.925\linewidth]{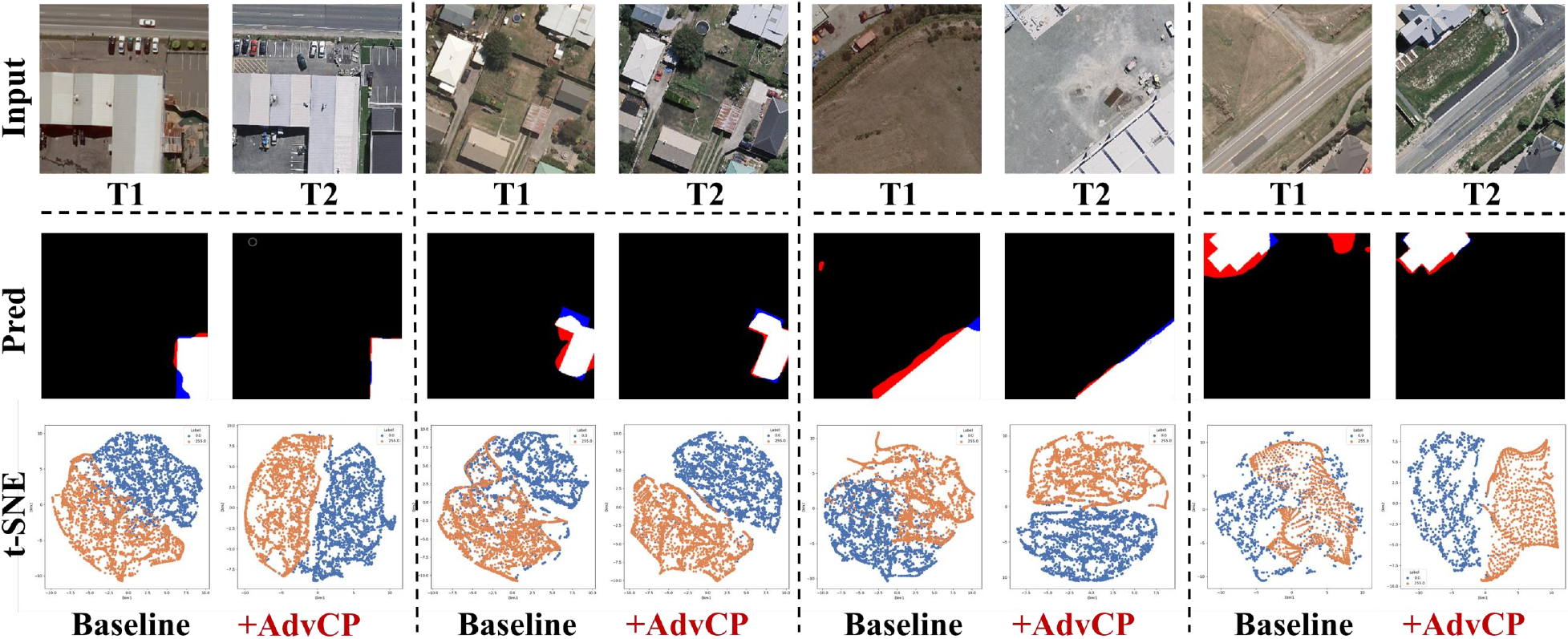}
  \caption{Visualization of feature distribution improvement. In the third row of the t-SNE images,  points represent unchanged samples, and orange points are changed samples. It is evident that unchanged and changed features become more distinguishable. In the second row, consistent improvements are observed in predictions. For clarity, in the predictions, \textcolor{red}{false positive (erroneous changed) pixels} are marked in red color, and \textcolor{blue}{false negative (erroneous unchanged) pixels} are marked in  color.} \label{experiment4}
\end{figure*}

To observe the impact of AdvCP on the distributions of changed and unchanged pixels in the latent space, we visualize the feature manifold before and after integrating AdvCP using t-SNE \cite{maaten2008visualizing}.

As shown in Fig. \ref{experiment4}, the overlapping unchanged and changed features become more distinct after integrating AdvCP. Additionally, the prediction results show improvements that align with the feature distribution. This analysis supports the argument that AdvCP helps the classifier to identify co-occurring background variation samples, thereby creating a more discriminative boundary between changed and unchanged features.

\begin{table}[ht]
\renewcommand{\arraystretch}{1.2}
\centering
\caption{Training time (minutes per 100 iterations) and GPU overhead (GB) of our AdvCP. All results are evaluated on the training split of the LEVIR-CD dataset.}
\label{time}\small
\begin{tabular}{ccc} 
\hline\hline
\textbf{Method}           & \textbf{Training Time} & \textbf{GPU Overhead}  \\ 
\hline
WCDNet                    & 2.69                   & 6.32                   \\
\textbf{Ours w/ WCDNet}   & \textbf{2.73}& \textbf{7.15}\\ 
\hline
TransWCD                  & 3.46                   & 4.52                   \\
\textbf{Ours w/ TransWCD} & \textbf{3.49}& \textbf{5.35}\\
\hline\hline
\end{tabular}
\end{table}

\subsection{Computational Complexity} \label{sec:complexity}
Our AdvCP method can be seamlessly integrated into the existing WSCD pipeline via a plug-and-play setup. As shown in Table \ref{time}, we assess the computational efficiency of AdvCP by comparing its training and inference times with the TransWCD baseline. Incorporating AdvCP results in a minimal increase of 0.04 and 0.03 minutes per 100 iterations for training on the two baselines, respectively. Furthermore, since AdvCP is used exclusively during the training phase, it does not introduce any additional overhead during inference. The impact on GPU memory is minimal, with an increase of only 0.83GB.

\begin{table}
\renewcommand{\arraystretch}{1.2}
\setlength{\tabcolsep}{4pt}
\centering
\caption{Evaluation of different prototype granularities for AdvCP.  F1 score (\%) are reported on  three datasets.}
\label{granularity}\small
\begin{tabular}{cccc} 
\hline\hline
\textbf{Calculation Granularity}& \textbf{WHU-CD}& \textbf{LEVIR-CD}& \textbf{DSIFN-CD}\\ 
\hline
Baseline (TransWCD)                        & 68.73& \textcolor{black}{60.08}& \textcolor{black}{53.41}  
\\ 
\hline
Image-wise& 69.13& \textcolor{black}{60.95}& \textcolor{black}{54.03}  
\\
Batch-wise& 71.52& \textcolor{black}{63.14}& \textcolor{black}{56.09}  
\\
Frozen Global& 70.66& \textcolor{black}{62.60}& \textcolor{black}{55.24}  
\\
\textbf{Online Global (Ours)}& \textbf{74.45}& \textcolor{black}{\textbf{66.03}}& \textcolor{black}{\textbf{58.67}}  \\
\hline\hline
\end{tabular}
\end{table}

\subsection{Ablation Study}
\noindent\textbf{Granularity of Clustering Prototypes.}\label{sec:granularity} We examine the impact of different granularities for AdvCP's clustering anchors. Granularities include prototypes specifically global, image-wise, and batch-wise prototypes. The global prototype has two calculation approaches. One is the frozen global prototype, which is directly extracted once from the entire training dataset at the initial computation. The other is our accumulated online global prototype that is calculated through each batch's exponentially weighted moving average. 

TransWCD is selected as the baseline in the ablation studies, and also applied in the following experiments. \textcolor{black}{As reported in Table \ref{granularity}, on all three datasets, image-wise prototypes provide minor improvements over the baseline, with +0.40\%, +0.87\%, and +0.62\% F1 score for WHU-CD, LEVIR-CD, and DSIFN-CD, respectively. Batch-wise prototypes show more substantial gains, achieving +2.79\%, +3.06\%, and +2.68\% F1 score. The frozen global prototype performs in between, with +1.93\%, +2.52\%, and +1.83\% F1 score, likely because it is computed only once at the beginning and cannot adapt to training.}

\textcolor{black}{Our online global prototype achieves the best performance across all datasets, with +5.72\%, +5.95\%, and +5.26\% F1 score improvements on WHU-CD, LEVIR-CD, and DSIFN-CD, respectively. By continuously updating the global feature center through an exponentially weighted moving average, it balances capturing global characteristics with adapting to batch-wise variations, thereby alleviating local random sampling biases (e.g., image-wise or batch-wise) and enhancing the reliability of model learning.}

\begin{table}
\setlength{\tabcolsep}{3.5pt}
\renewcommand{\arraystretch}{1.2}
\centering
\caption{Improvement of AdvCP integrating with different loss functions. F1 score (\%) and IoU (\%) are reported.}
\label{AdvCP_vary_loss}\small
\begin{tabular}{ccccccc} 
\hline\hline
\multirow{2}{*}{\textbf{Method}} & \multicolumn{2}{c}{\textbf{WHU-CD}}& \multicolumn{2}{c}{\textbf{LEVIR-CD}} & \multicolumn{2}{c}{\textcolor{black}{\textbf{DSIFN-CD}}
}\\ 
\cline{2-7}& \textbf{F1} & \textbf{IoU} & \textbf{\textbf{F1}} & \textbf{\textbf{IoU}}    & \textcolor{black}{\textbf{F1}}&\textcolor{black}{\textbf{IoU}}\\ 
\hline
Baseline (TransWCD)                           & 68.73         & 52.36          & 60.08                  & 42.94                      & \textcolor{black}{53.41}&\textcolor{black}{36.44}\\
\hline
Consistency                     & 72.79         & 56.71          & 63.99                  & 45.34                      & \textcolor{black}{56.16}&\textcolor{black}{39.59}\\
Contrastive                     & 70.99         & 55.99          & 64.08                  & 48.13                      & \textcolor{black}{55.40}&\textcolor{black}{38.26}\\
\textbf{Center-Accumulated}& \textbf{74.45}& \textbf{59.30}& \textbf{66.03}& \textbf{50.40}& \textcolor{black}{\textbf{58.67}}&\textcolor{black}{\textbf{42.14}}\\
\hline\hline
\end{tabular}
\end{table}

\noindent\textbf{Different Constraints for AdvCP.} \label{sec:diff_loss} 
We also compare the proposed AdvCP (i.e., center-accumulated) loss $\mathcal{L}_{adv}$ with other optimization objective functions designed to enhance pseudo-label quality. Specifically, we consider consistency loss \cite{lee2022weakly} and contrastive loss \cite{du2022weakly}. In our implementation, consistency loss is extended to a pixel-wise version, promoting consistent distribution across samples. The performance of AdvCP integrated with different losses is shown in Table \ref{AdvCP_vary_loss}.

On the WHU-CD dataset, applying consistency loss, contrastive loss, and our AdvCP loss results in F1 score improvements of +4.06\%, +2.26\%, and +5.72\% over the TransWCD baseline, respectively. \textcolor{black}{Similar trends are observed on the LEVIR-CD and DSIFN-CD datasets, with our AdvCP loss showing peak increases of +5.95\% and +5.26\% on F1 score, respectively. Our center-accumulated loss consistently outperforms consistency and contrastive losses, demonstrating its superiority. We argue that this effectiveness arises from its ability to keep clustering centers stable over time. By gradually updating global information, prototypes are protected from local batch fluctuations, making the model more robust in handling co-occurring noise.}

\noindent\textbf{Momentum Coefficient $\boldsymbol{\lambda}$.} 
\textcolor{black}{Table \ref{img_inst} reports the effect of different momentum coefficient $\lambda$ settings on the AdvCP loss (Eq. \ref{ctr}) across three datasets. The coefficient $\lambda$ controls the update rate by balancing contributions from historical and current batch data.}

We observe that $\lambda=0.5$, which balances the contributions from the current and previously accumulated feature centers, consistently achieves the best performance across all three datasets. Specifically, it yields F1/IoU of 74.45\%/59.30\%, 66.03\%/50.40\%, and 58.67\%/42.14\% on the WHU-CD, LEVIR-CD, and DSIFN-CD datasets, respectively. 

\textcolor{black}{This trend reflects a clear trade-off principle. When $\lambda$ is too small (e.g., 0.0), the global centers become overly dependent on the current batch, leading to instability. In contrast, a very large $\lambda$ (e.g., 1.0) slows down convergence by ignoring useful batch-specific information. Selecting $\lambda=0.5$ provides a balanced solution in which stability, convergence speed, and generalization ability are jointly optimized. Moreover, the incorporation of accumulated centers effectively mitigates sampling bias across individual batches.}

\begin{table}
\setlength{\tabcolsep}{3.5pt}
\renewcommand{\arraystretch}{1.2}
\centering
\caption{Comparison of different momentum coefficient $\lambda$ for the AdvCP loss.  F1 score (\%) and IoU (\%) are reported.}
\label{img_inst}\small
\begin{tabular}{ccccccc} 
\hline\hline
\multirow{2}{*}{\textbf{$\boldsymbol{\lambda}$}}& \multicolumn{2}{c}{\textbf{WHU-CD}}& \multicolumn{2}{c}{\textcolor{black}{\textbf{LEVIR-CD}}}& \multicolumn{2}{c}{\textcolor{black}{\textbf{DSIFN-CD}}}\\
 \cline{2-7}&  \textbf{F1}&\textbf{IoU}                &  \textcolor{black}{\textbf{F1}}&\textcolor{black}{\textbf{IoU}}& \textcolor{black}{\textbf{F1}}&\textcolor{black}{\textbf{IoU}}\\ 
\hline
Baseline (TransWCD) & 68.73 & 52.36 & \textcolor{black}{60.08} & \textcolor{black}{42.94} & \textcolor{black}{53.41} & \textcolor{black}{36.44}\\ 
\hline
0.0 & 72.98 & 57.06 & \textcolor{black}{64.06} & \textcolor{black}{48.83} & \textcolor{black}{57.24} & \textcolor{black}{41.62}\\
0.1 & 73.69 & 58.71 & \textcolor{black}{64.87} & \textcolor{black}{49.21} & \textcolor{black}{57.88} & \textcolor{black}{41.98}\\
0.5 & \textbf{74.45} & \textbf{59.30} & \textcolor{black}{\textbf{66.03}} & \textcolor{black}{\textbf{50.40}} & \textcolor{black}{\textbf{58.67}} & \textcolor{black}{\textbf{42.14}}\\
0.7 & \multicolumn{1}{c}{72.48} & 56.68 & \multicolumn{1}{c}{\textcolor{black}{63.95}} & \textcolor{black}{47.96} & \multicolumn{1}{c}{\textcolor{black}{56.95}} & \textcolor{black}{40.85}\\
1.0 & 71.52 & 54.47 & \textcolor{black}{63.14} & \textcolor{black}{47.11} & \textcolor{black}{56.09} & \textcolor{black}{38.79}\\
\hline\hline
\end{tabular}
\end{table}
\begin{figure*}
	\centering
	\includegraphics[scale=0.55]{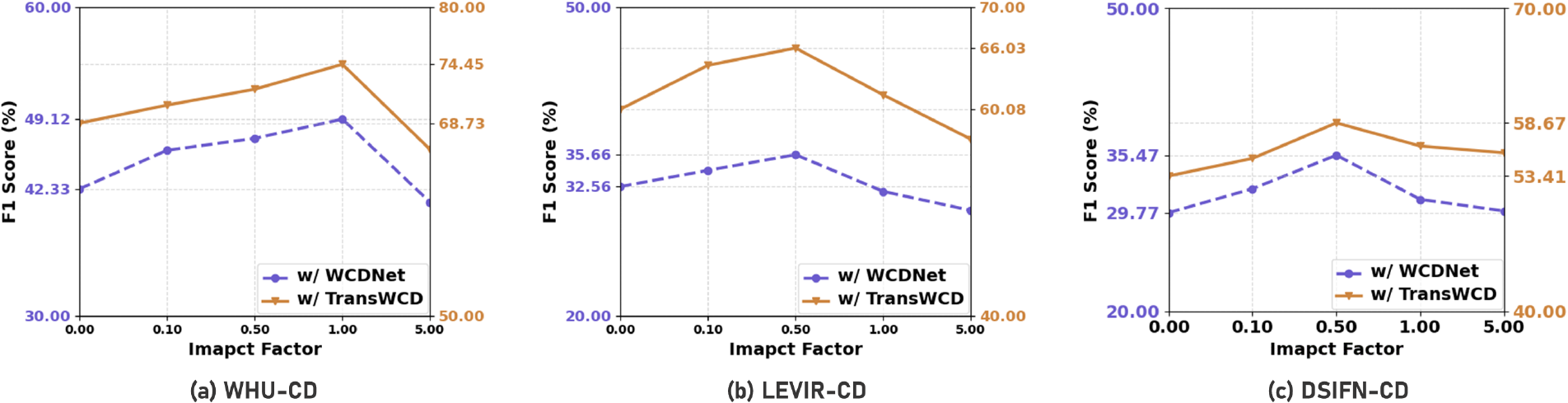}
	\caption{Impact factor $\alpha$ for the AdvCP loss $\mathcal{L}_{adv}$. F1 score (\%) is reported on the WHU-CD, LEVIR-CD, and DSIFN-CD datasets.}
	\label{appendix1}
\end{figure*}
\noindent\textbf{Impact Factor \textbf{$\alpha$}.} \textcolor{black}{In Fig. \ref{appendix1}, we conduct empirical experiments to check the impact of varying $\alpha$, which controls the contribution of the AdvCP loss $\mathcal{L}_{adv}$ to the overall loss, as defined in Eq. \ref{total_loss}}. When we adjust the impact factor $\alpha$ on top of two baseline methods, WCDNet and TransWCD, we find that the overall optimal choice of $\alpha$ on the WHU-CD, and LEVIR-CD, and DSIFN-CD datasets are 1.00, 0.50, and 0.50, respectively. On the WHU-CD dataset, WCDNet rises by 6.79\%, and TransWCD reaches its highest 74.45\% F1 score at the optimal impact factor with $\alpha=1.00$. 

While, on the LEVIR-CD dataset, the impact factor $\alpha=0.50$ brings a gain of 5.95\% F1 score. Before reaching these optimal points, a consistent positive correlation between performance and the parameter is observed. Beyond these optimal points, further increasing the impact factor $\alpha$ may lead to decreased results.

This decline is likely due to an overemphasis on co-occurring unchanged background pixels, thus disrupting the optimization of changed pixels \cite{10680274}. The DSIFN-CD dataset parallels the LEVIR-CD in terms of performance.

\section{Study on Extensive Tasks}
\begin{table}[ht]
\setlength{\tabcolsep}{20pt}
\renewcommand{\arraystretch}{1.2}
\centering
\caption{Performance of the AdvCP extension for weakly-supervised multi-class dense prediction on the COCO and PASCAL VOC datasets, reported as mIoU (\%).}
\label{wsss}\small
\begin{tabular}{ccc} 
\hline\hline
\textbf{Method} & \textbf{COCO} & \textbf{VOC} \\ 
\hline
SEAM \cite{wang2020self}          & 31.90  & 64.50 \\
\textbf{+ AdvCP (Ours)}             & \textbf{33.74} & \textbf{65.83} \\ 
\hline
MCTformer \cite{xu2022multi}      & 42.00  & 71.90 \\
\textbf{+ AdvCP (Ours)}             & \textbf{44.16} & \textbf{72.34} \\ 
\hline\hline
\end{tabular}
\end{table}

\subsection{Extending AdvCP to Multi-Label Weakly-Supervised Dense Prediction}
\noindent\textbf{Multi-Label AdvCP. } In this section, we further extend the AdvCP framework from binary-label weakly-supervised change detection to multi-label dense prediction tasks. This extension addresses the common challenge of co-occurring error-responsive pixels present in multi-label scenarios, where single images contain pixels belonging to multiple classes.

For multi-label tasks, each image example comprises multi-class pixels. The one-hot multi-class labels, represented as vectors \([y_1, y_2, \ldots, y_K]\) with \(y_i \in \{0,1\}\) for \(K\) categories, can be viewed as a combination of multiple binary-class labels. In the multi-label AdvCP framework, we transform each one-hot vector into an all-ones vector as input. After this transformation, we generate a multi-class AdvCP sample mask \(\mathcal{M}_{adv_K} \in \mathbb{R}^{N \times HW \times K}\) using the XOR operation between the all-ones vector response \(\mathcal{C}_{all_1}\) and the original response \(\mathcal{C}\):
\begin{equation}
\mathcal{M}_{adv_K} = \mathcal{C}_{all_1} \oplus \mathcal{C}.
\end{equation}
Here, \(\mathcal{M}_{adv_K}\) contains the detected adversarial pixel samples for \(K\) categories, which are erroneous predictions that should not exist in the current images.

To rectify these erroneously predicted samples, we perform clustering towards the accumulated global feature centers of the relevant categories. This process aligns the features of the mispredicted pixels with the correct category centers, effectively reducing the impact of co-occurring errors. The corresponding multi-class AdvCP loss \(\mathcal{L}_{adv_k}\) is defined as:
\begin{equation} 
\mathcal{L}_{adv_K} = \sum^K_{k=1} \| \mathcal{F}_{adv_k} - p_{ctr_k} \|_2^2,
\end{equation}
where \(\mathcal{F}_{adv_k}\) represents the feature embeddings of the AdvCP samples, and \(p_{ctr_k}\) denotes the global feature center for the category \(k\).

\begin{figure}[t]
	\centering
	\includegraphics[width=1.0\linewidth]{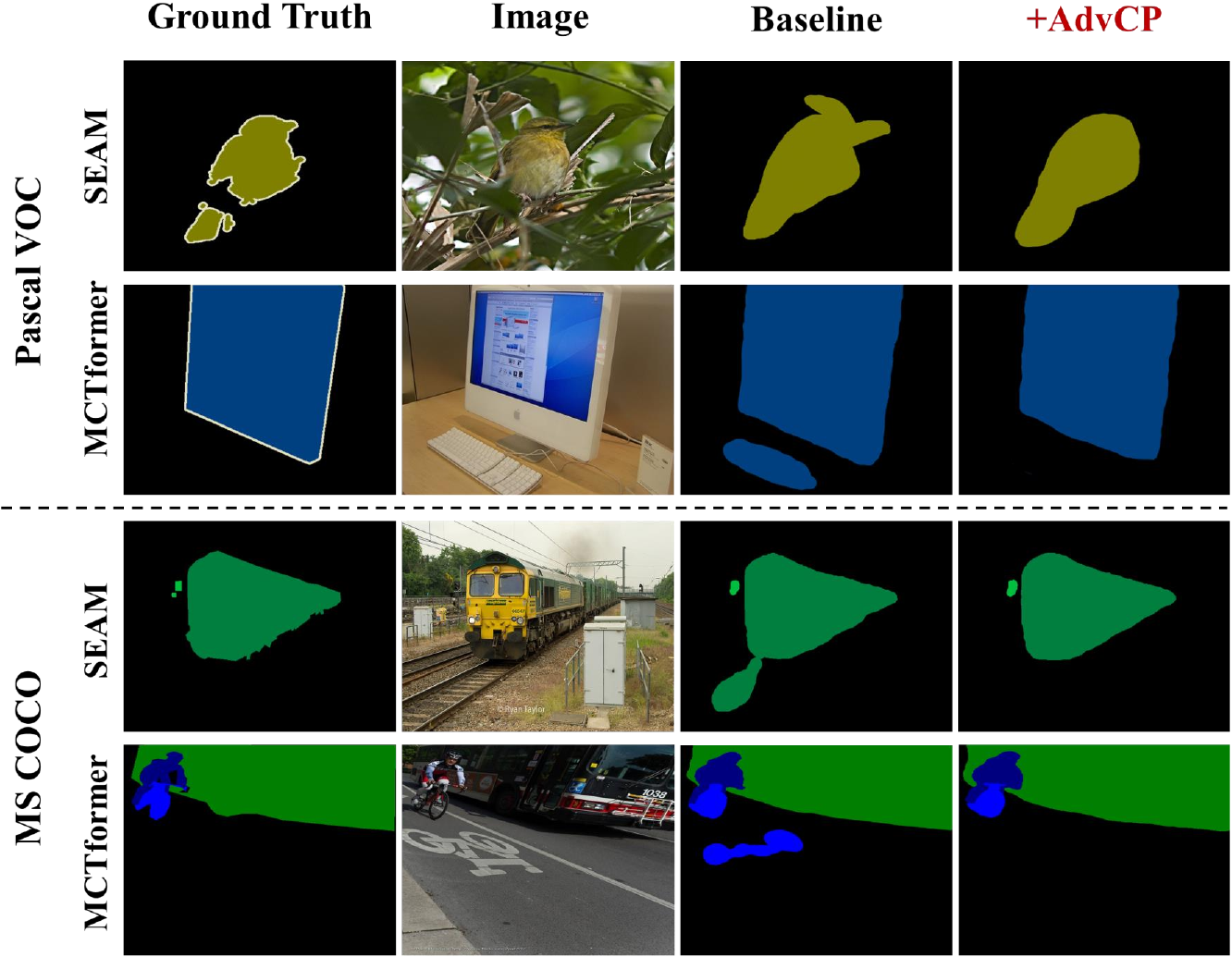}
	\caption{Qualitative improvements of the proposed AdvCP on the multi-class {MSCOCO and PASCAL VOC} datasets.}
	\label{experiment5}
\end{figure}

\noindent\textbf{Multi-Label Experiments.} We evaluate the multi-class AdvCP framework on the MSCOCO and PASCAL VOC 2012 datasets, as shown in Table \ref{wsss} and visualized in Fig. \ref{experiment5}. The proposed AdvCP method enhances the performance of weakly-supervised semantic segmentation models, specifically SEAM \cite{wang2020self} and MCTformer \cite{xu2022multi}, achieving mIoU improvements of +1.84\% and +2.16\% on the MSCOCO dataset, and +1.33\% and +0.44\% on the PASCAL VOC dataset, respectively. Moreover, the visualized results show that the co-occurring noisy pixels have been removed. For example, in the second row, the erroneous keyboard, predicted as part of the monitor, is corrected. These experimental results demonstrate the effectiveness of our AdvCP method in multi-class weakly-supervised dense prediction.

\subsection{Extending AdvCP to Fully-Supervised Change Detection}
\noindent\textbf{Fully-Supervised AdvCP. }We extend the scope of AdvCP from image-level weakly-supervised change detection (WSCD) to pixel-level fully-supervised change detection (FSCD). We assume the pixel-level binary labels are denoted as \( Y \in \mathbb{R}^{N \times HW} \), where each pixel is labeled as either `unchanged' or `changed'. In the FSCD setting, the adversarial prompt perturbations are introduced by uniformly assigning the pixel-level `changed' labels to all pixels after each batch of training. This reassignment forces the model to respond to `changed' in every pixel location of `unchanged' images.

After the label perturbation, we calculate the weights \( w_c \) based on the pixel-level `changed' perturbed labels, which are used to obtain the all-change localization maps \( \mathcal{C}_c \) in Eq.~\ref{c_c}. Specifically, the change detection model utilizes the gradients of the feature maps \( \mathcal{F} \in \mathbb{R}^{N \times HW \times D} \) with respect to the predicted pixel-level `changed' probability scores \( \hat{Y} \in \mathbb{R}^{N \times HW} \):

\begin{equation}
\label{weight_calculation}
w_c(h,w) = \frac{1}{HW} \sum_{i=1}^D \frac{\partial \mathcal{F}^{(i)}(h,w)}{\partial \hat{Y}(h,w)}.
\end{equation}
The remaining operations of AdvCP for the FSCD setting are the same as those for WSCD.

The final loss \( \mathcal{L} \) in the FSCD setting consists of the binary cross-entropy pixel-level classification loss \( \mathcal{L}_{\text{pixel}} \) and the AdvCP loss \( \mathcal{L}_{\text{adv}} \):
\begin{equation}
\mathcal{L} = \mathcal{L}_{\text{pixel}} + \alpha \mathcal{L}_{\text{adv}},
\end{equation}
where the impact factor is empirically set to \( \alpha = 0.1 \).

\begin{table}[!ht]
 \setlength{\tabcolsep}{2.3pt}
\renewcommand{\arraystretch}{1.2}
\centering
\caption{Quantitative improvements over fully-supervised change detection methods. F1 score (\%) and IoU (\%) are reported.}
\label{FSCD}\small
\begin{tabular}{clcclcclccl} 
\hline\hline
\multirow{2}{*}{\textbf{Method}}                       &  & \multicolumn{2}{c}{\textbf{WHU-CD}}& \multicolumn{1}{c}{} & \multicolumn{2}{c}{\textbf{\textbf{LEVIR-CD}}}&  & \multicolumn{2}{c}{\textbf{\textbf{DSIFN-CD}}}& \multicolumn{1}{c}{}  \\ 
\cline{2-11}&  & \textbf{F1}                          & \textbf{IoU}                         &                      & \textbf{F1}                          & \textbf{\textbf{IoU}}                &  & \textbf{F1}                          & \textbf{\textbf{IoU}}                &                       \\ 
\hline
BIT&  & 86.64& 83.98&                      & 89.24& 80.68&  & 69.26& 52.97&                       \\
\textbf{+AdvCP (Ours)}&  & \textbf{87.78}& \textbf{85.02}&                      & \textbf{90.57}& \textbf{81.75}&  & \textbf{71.86}& \textbf{56.04}&                       \\
\hline
\multicolumn{1}{c}{Changer}                                   &  & \multicolumn{1}{c}{88.15}                                  & \multicolumn{1}{c}{85.05}                 &                      & \multicolumn{1}{c}{92.26}                                  & \multicolumn{1}{c}{82.73}                 &  & \multicolumn{1}{c}{70.94}                                  & \multicolumn{1}{c}{58.30}                 &                       \\
\textbf{+AdvCP (Ours)}&  & \textbf{89.02}& \textbf{86.21}&                      & \textbf{93.11}& \textbf{83.59}&  & \textbf{72.67}& \textbf{60.05}&                       \\
\hline\hline
\end{tabular}
\end{table}

\noindent\textbf{Fully-Supervised Experiments.} Experiments demonstrate the effectiveness of the proposed AdvCP method over existing change detection methods. We leverage OpenCD, the open-source toolkit for change detection~\cite{li2024open}, using BIT~\cite{chen2021remote} and Changer~\cite{fang2023changer} as baseline models. BIT is a CNN-Transformer model that abstracts pixel-level features into tokens to capture spatio-temporal context. Changer interacts with bi-temporal features using aggregation-distribution and feature exchange strategies. The experimental results on the WHU-CD, LEVIR-CD, and DSIFN-CD datasets are presented in Table \ref{FSCD}.

On the WHU-CD dataset, the BIT model’s F1 score improves from 86.64\% to 87.78\%, and the IoU rises from 83.98\% to 85.02\%, marking increases of +1.14\% and +1.04\%, respectively. Moving to the more challenging LEVIR-CD dataset, the BIT model again benefits from AdvCP, with the F1 score climbing from 89.24\% to 90.57\%, an increase of +1.33\%, and the IoU rising from 80.68\% to 81.75\%, a gain of +1.07\%.

The most significant improvements are observed on the challenging DSIFN-CD dataset, where background variations present more complexity. The BIT model's F1 score increases from 69.26\% to 71.86\%, and the IoU rises from 52.97\% to 56.04\%, yielding substantial gains of +2.60\% in F1 and +3.07\% in IoU. Similarly, the Changer model demonstrates notable progress on this dataset, with the F1 score rising from 70.94\% to 72.67\%, an increase of +1.73\%, and the IoU improving from 58.30\% to 60.05\%, a gain of +1.75\%.

We attribute the effectiveness of AdvCP in fully-supervised settings to its enhanced ability to focus on hard samples, as described in Section \ref{sec:method3}. This focused attention enables AdvCP to handle challenging cases, resulting in more robust and accurate change detection \cite{10599227}. Preliminary experiments demonstrate the generalization and potential of our AdvCP framework, with further evaluations planned for broader applicability.

\section{Conclusion}
In this paper, we address the challenge of co-occurring background variation noise in Weakly-Supervised Change Detection (WSCD). Instead of employing complex methods to directly detect these co-occurring change pixels under weak supervision, we identify them by analyzing unchanged image pairs. To achieve this, we propose Adversarial Class Prompting (AdvCP). AdvCP operates by inverting image-level labels from unchanged to changed, thereby inducing adversarial responses in unchanged scenes. We then design an AdvCP loss function that directs the clustering of adversarial pixels toward an online global prototype, derived from all training data using an exponentially weighted moving average. We integrate AdvCP with six representative baseline methods; our experiments show that AdvCP significantly improves their performance across various change detection scenarios. Additionally, we extend AdvCP to fully-supervised change detection and multi-class weakly-supervised dense prediction tasks, demonstrating its flexibility and versatility.


\bibliographystyle{IEEEtran}
\bibliography{reference}

\end{document}